\documentclass[10pt,twocolumn,letterpaper]{article}

\usepackage[pagenumbers]{cvpr} 

\usepackage{times}
\usepackage{epsfig}
\usepackage{graphicx}
\usepackage{amsmath}
\usepackage{amssymb}

\usepackage{bbm}
\usepackage{algorithm}
\usepackage{algorithmic}
\usepackage{subcaption}
\usepackage{booktabs}
\usepackage{multirow}
\usepackage{nicefrac}

\usepackage{bm}

\usepackage{xcolor}
\definecolor{myblue}{RGB}{46, 117, 182}
\definecolor{myred}{RGB}{237, 125, 49}
\definecolor{myyellow}{RGB}{255, 192, 0}
\definecolor{mygreen}{RGB}{0, 176, 80}

\usepackage{gensymb}
\newcommand\mdoubleplus{\mathbin{+\mkern-10mu+}}

\definecolor{cvprblue}{rgb}{0.21,0.49,0.74}
\usepackage[pagebackref,breaklinks,colorlinks,citecolor=cvprblue]{hyperref}

\def\paperID{2673} 
\def\confName{CVPR}
\def\confYear{2024}

\title{Optimizing Camera Configurations for Multi-View Pedestrian Detection}

\author{%
Yunzhong Hou\textsuperscript{$\dagger$} 
\and 
Xingjian Leng\textsuperscript{$\dagger$}
\and 
Tom Gedeon\textsuperscript{$\ddagger$}
\and 
Liang Zheng\textsuperscript{$\dagger$}
\and
\textsuperscript{$\dagger$} Australian National University \\
\texttt{ \small \{firstname.lastname\}@anu.edu.au}
\and 
\textsuperscript{$\ddagger$} Curtin University \\ 
\texttt{ \small \{firstname.lastname\}@curtin.edu.au}
}

\begin{document}
\maketitle
\begin{abstract}


Jointly considering multiple camera views (multi-view) is very effective for pedestrian detection under occlusion. For such multi-view systems, it is critical to have well-designed camera configurations, including camera locations, directions, and fields-of-view (FoVs). Usually, these configurations are crafted based on human experience or heuristics. In this work, we present a novel solution that features a transformer-based camera configuration generator. Using reinforcement learning, this generator autonomously explores vast combinations within the action space and searches for configurations that give the highest detection accuracy according to the training dataset. The generator learns advanced techniques like maximizing coverage, minimizing occlusion, and promoting collaboration. Across multiple simulation scenarios, the configurations generated by our transformer-based model consistently outperform random search, heuristic-based methods, and configurations designed by human experts, shedding light on future camera layout optimization. 
\end{abstract}

\section{Introduction}
\label{sec:intro}

Multiple camera views (multi-view) are becoming more and more popular for machine understanding. Unlike a monocular view (only one camera), multi-view systems can jointly consider different cameras at different angles to address challenges in recognition including ambiguities, occlusions, and limited field-of-view (FoV) coverage. To mention a few examples, multi-view classification \cite{su2015multi,qi2016volumetric} uses circular or spherical camera rigs to capture an object from diverse angles and reduce ambiguities; multi-view stereo \cite{seitz2006comparison,yao2018mvsnet} recovers the 3D shape from a collection of 2D images; autonomous vehicles \cite{caesar2020nuscenes,huang2021bevdet,li2022bevformer} and multi-camera tracking \cite{ristani2016performance,hou2021adaptive} expands the FoV coverage with multiple cameras. In this work, we focus on multi-view pedestrian detection \cite{chavdarova2018wildtrack,hou2020multiview}, a specific usage of multi-view systems that use multiple cameras to jointly combat occlusions, reduce ambiguities, increase FoV coverage, and ultimately locate pedestrians in 3D from a bird's-eye-view (BEV).

\begin{figure}
\centering
    \begin{subfigure}[b]{\linewidth}
    \centering
        \includegraphics[width=\linewidth]{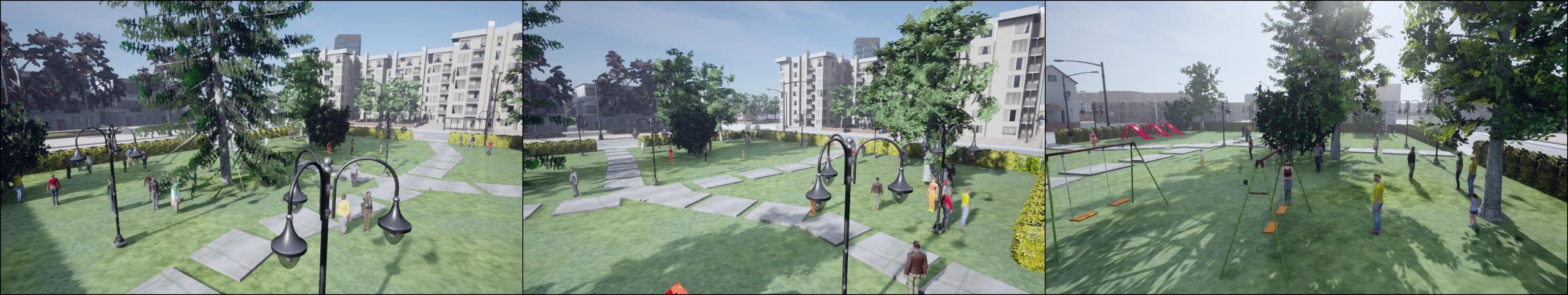}
        \includegraphics[width=\linewidth]{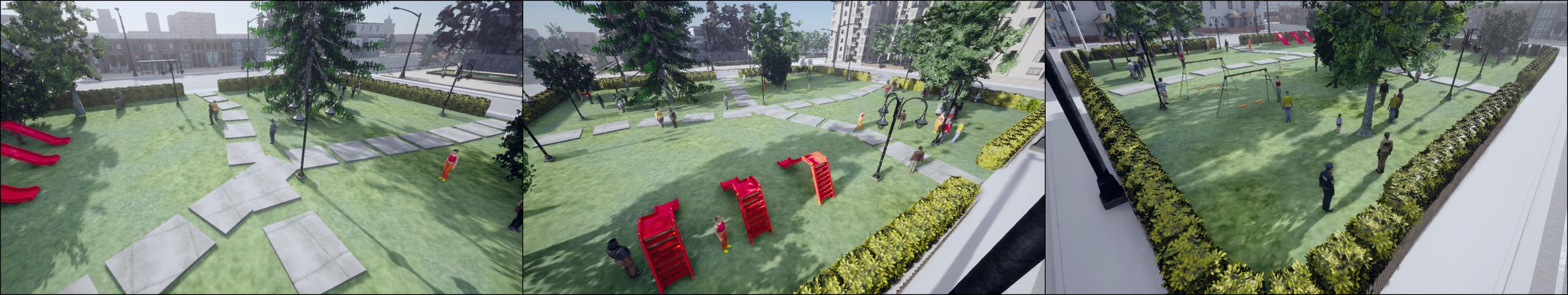}
        \caption{Camera views given by a human expert (top) and our method (bottom).}
    \end{subfigure}
    \begin{subfigure}[b]{\linewidth}
    \centering
        \includegraphics[width=0.45\linewidth]{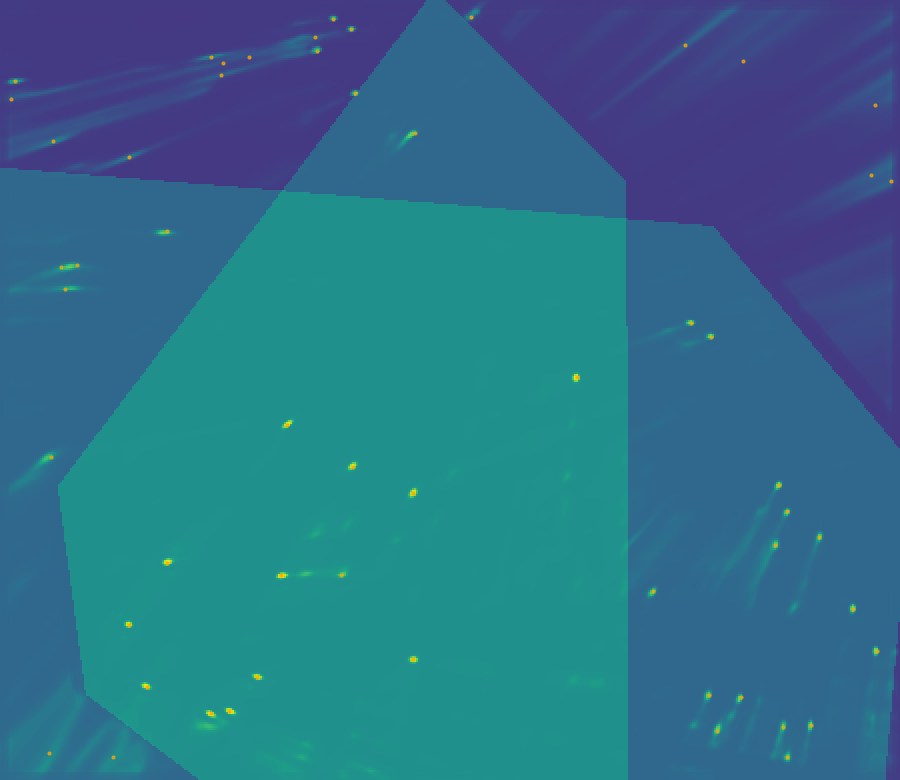}
        \hfill
        \includegraphics[width=0.45\linewidth]{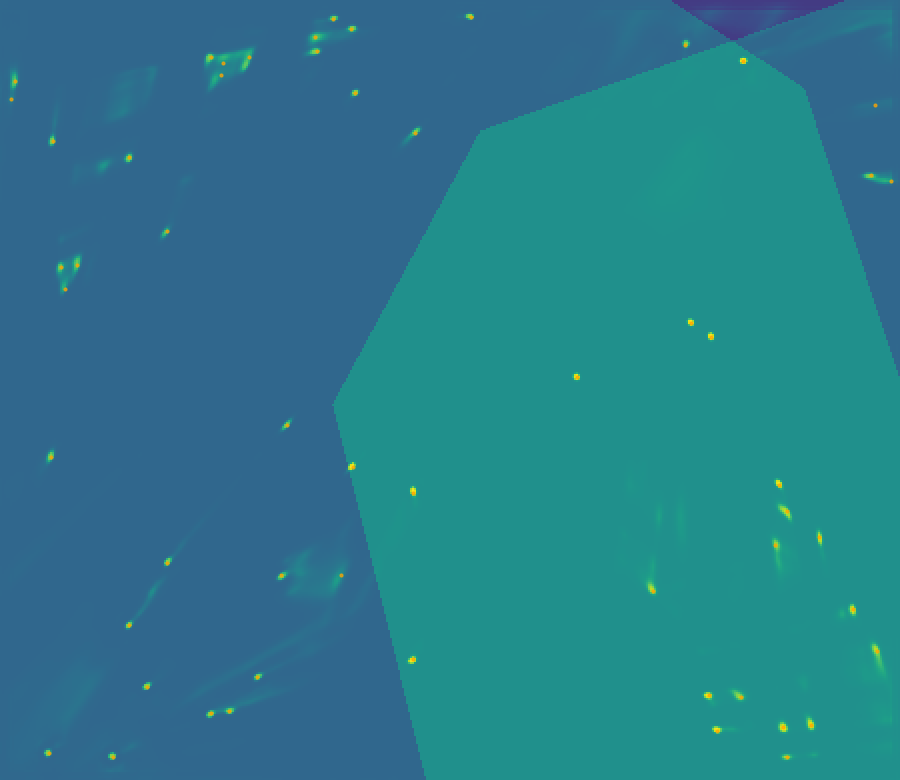}
        \caption{FoV coverage from a human expert (left) vs. our search method (right). Less blue colors indicate the area being covered by more cameras. Zoom in to see the ground truth pedestrian locations (orange dots) and detection results (heatmap overlay).}
    \end{subfigure}
\caption{
Comparison between human expert design and searched configuration. On the test set of this scenario, these two configurations achieve 69.9\% and 91.9\% MODA, respectively. 
}
\label{fig:demo}
\end{figure}

For these multi-view systems,  camera configurations including locations, directions, and FoVs are critical parameters. Poor camera configurations can impact overall system performance and are very difficult to remedy with image processing tools like CNNs, since the content in images may already be limited. 

Usually, camera configurations are determined by {human experts} or based on {heuristics}. On the one hand, \underline{\textit{experts}} design the configurations based on experience \cite{zong2018method} or general guidelines \cite{hori1997traffic,vigderman_2023_where,conceptdraw_camera}. Unfortunately, these designs are usually directly adopted without further validation. 
On the other hand, \underline{\textit{heuristic-based methods}} focus on the FoV coverage and study camera placements that ensure each location on building floor plans is covered by at least one camera \cite{erdem2004optimal,erdem2006automated}. However, these methods do not consider dynamic occlusions from foreground objects such as pedestrians or vehicles, and more importantly, the image processing step (\eg, object detection).
In fact, neither of these existing solutions can guarantee the overall system performance for their configurations. 

In this paper, we aim to directly optimize the camera configurations with the final multi-view pedestrian detection accuracy as the objective. Unlike previous approaches that either rely on human experience \cite{zong2018method,hori1997traffic} or the heuristic of FoV coverage \cite{erdem2006automated}, the proposed optimization is directed by quantitative analysis, while taking both dynamic occlusions and the pedestrian detection network into account. In Fig.~\ref{fig:demo}, we demonstrate some human expert designs and compare them with our searched configurations. Compared to the human expert configurations, the searched results are more creative with the camera positioning, and can find views with less obstruction yet more collaboration and overlapping FoV. Thus, even with the same pedestrian detector, the proposed configuration provides more advantages than expert settings. 

To study this problem, we formulate the camera configuration search as an interactive process between an \textbf{agent} who can predict camera configurations, and an \textbf{environment} that takes the predicted configurations and returns the pedestrian detection accuracy as their efficacy. 
Specifically, we first build the \underline{\textit{interactive environment}}. By connecting the Carla \cite{dosovitskiy2017carla} simulation engine for autonomous vehicles and existing multi-view pedestrian detection networks, we introduce CarlaX, an interactive playground that can render camera views from given configurations and return the detection accuracy of specific multi-view pedestrian detectors as feedback. 
We then introduce a novel \underline{\textit{camera configuration generator}} as the agent. Using a transformer sequence predictor, the network generates the location, direction, and FoV for the next camera, iteratively. 
Importantly, to navigate through the non-differentiable parts\footnote{Interaction with the simulation engine, rendering, \etc} while sufficiently exploring the vast combinations\footnote{For a $20\times 20$ m$^2$ square with camera height from 0 to 4 m, quantizing the spatial resolution by 0.1 m would give 1.6 million combinations. Note that camera rotations and FoV are yet to be factored in, and we have to do this for all $N$ cameras.}, we introduce a reinforcement-learning-based \underline{\textit{training scheme}} to effectively search the configuration space while jointly training the multi-view detection network. We also introduce two differentiable regularization terms to make the reinforcement learning search process more effective. 

Across multiple scenarios, our approach proves consistently advantageous over random search,  heuristic-based methods, and human expert design. We will make the CarlaX playground and the code for our approach publically available to promote future work on smart vision, and to incentivize the community to include more into optimization aside from neural network parameters.

Overall, our contributions are as listed follows,
\begin{itemize}
    \item a simulation environment for camera configuration study,
    \item a transformer-based camera configuration generator, and
    \item a reinforcement-learning-based training scheme that supports joint training for the detection network.
\end{itemize}

\section{Related Work}
\label{sec:related work}

\textbf{Multi-view pedestrian detection} aims to detect pedestrians from a bird’s-eye-view (BEV) under heavy occlusion, by jointly considering multiple camera views. 
Some \cite{chavdarova2017deep,baque2017deep} merge the multi-view anchor box feature to provide the overall scenario description. Others  \cite{hou2020multiview,song2021stacked,hou2021multiview} use the anchor-free approach and directly project image feature maps to BEV using camera tomography. 
In this work, we take MVDet architecture \cite{hou2020multiview} as the multi-view detector.

\textbf{Camera placement optimization}. 
A closely related line of work in computational geometry is the Art Gallary Problem \cite{o1987art,wikipedia_2022_art}, which aims to \underline{\textit{maximize the FoV coverage}}, and find the optimal camera locations that ensure each point on the building floor plan is visible and tries to minimize the number of cameras required \cite{erdem2004optimal,erdem2006automated}. Specifically, Sun \etal \cite{sun2021learning} investigate a learning-based camera placement method. 
However, in our study, we find there are many solutions that satisfy maximal coverage, but unlike our method, that directly optimizes for detection accuracy, such solutions often fail to produce the best detection results in complex scenes with occluded targets. 

Some also try to optimize the camera placements for \underline{\textit{better task performance}}. 
MVTN \cite{hamdi2021mvtn} uses a 3D point cloud as initial input and then estimates the best camera layout for each different object. 
Recently, Hou \etal \cite{hou2023learning} investigate how to select the most helpful views in multi-view systems for the best efficiency. 
A concurrent work by Klinghoffer \etal \cite{klinghoffer2023diser} also investigates camera placement. However, their experiments on stereo depth estimation and autonomous vehicles are at much smaller scales ($<$10 m$^3$) compared to this work ($>$1000 m$^3$). 

For \underline{\textit{3D reconstruction}}, researchers in active vision study a very similar problem of view planning where the location of 3D sensors are determined for 3D reconstructions \cite{chen2011active,fan2016automated}. 
View planning is also investigated in UAV reconstruction to maximize the constructability while minimizing the flight time \cite{zhou2020offsite,liu2022learning}. Lee \etal \cite{lee2022uncertainty} focus on NeRF \cite{mildenhall2021nerf} and find the best views to improve a coarse model. 
However, such methods often require hundreds of images to fully reconstruct a scenario and usually reconstruct static objects or buildings, whereas multi-view pedestrian detection usually jointly considers fewer than ten cameras due to GPU hardware constraints and focuses on moving pedestrians.

\begin{figure*}[ht]
\centering
\includegraphics[width=0.8\linewidth]{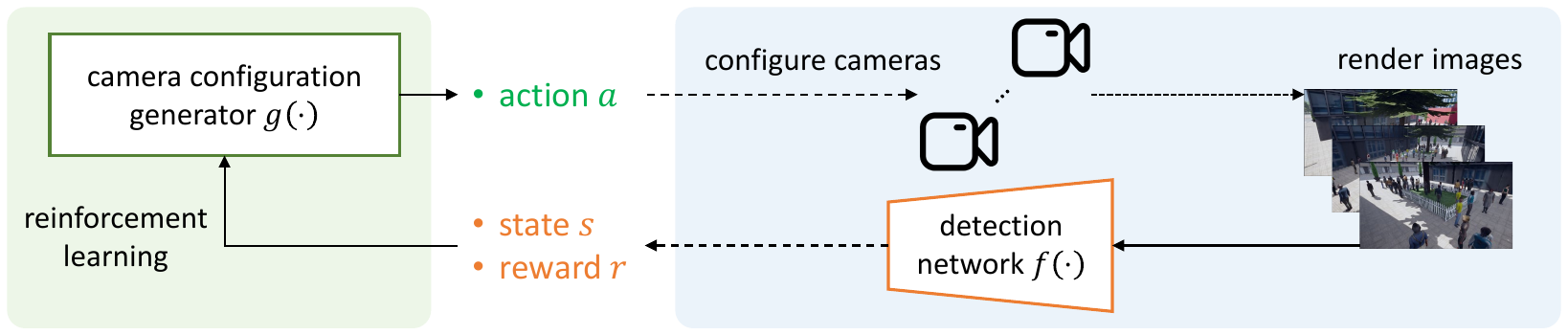}
\caption{
Overview of the proposed optimization method for camera configuration. 
On the left, the agent (generator) predicts the action $a$, \ie, the configuration for the next camera (Eq.~\ref{eq:config} and Eq.~\ref{eq:action}). On the right, the environment (simulation engine and detector) takes the predicted camera configurations and feeds rendered images to the multi-view pedestrian detector. It returns the list of configurations for existing cameras as the state $s$ (Eq.~\ref{eq:state}), and the detection accuracy as the reward $r$ (Eq.~\ref{eq:reward}). 
Dotted lines denote non-differentiable operations. 
}
\label{fig:overview}
\end{figure*}

\textbf{Other study on camera hardware.}
Over the years, researchers include more components in the imaging process into optimization, including sensor multiplexing for image quality \cite{chakrabarti2016learning}, phase-coded aperture \cite{haim2018depth} and focal length \cite{he2018learning} for single image depth estimation, structured illumination for active depth estimation \cite{baek2021polka}, point spread function (response to a point source of light) \cite{chang2019deep,tseng2021differentiable} LiDAR pulse configuration \cite{goudreault2023lidar} for detection, and optic components for HDR imaging \cite{sitzmann2018end,metzler2020deep}. 


\textbf{Reinforcement learning} (RL) guides an agent to maximize cumulative rewards while interacting with the environment. RL is usually modelled as a Markovian Decision Process (MDP), characterized by the state $s$, the action $a$, the internal transition $P_a(s,s')$ from state $s$ to state $s'$ after taking action $a$, and the reward $r$. The goal of reinforcement learning is to learn a policy $\pi\left(a|s\right)$ that maximizes the expected cumulative reward. To learn the best policy, value-based methods like Q-learning and DQN \cite{mnih2013playing} optimize the action value function $Q\left(s, a\right)$, which describes the estimated future return for a specific action $a$ at state $s$. Policy gradient methods like REINFORCE \cite{williams1992simple} and PPO \cite{schulman2017proximal} directly optimize for the polity $\pi\left(a|s\right)$. 
RL has seen a wide adoption in many computer vision tasks. Specifically, researchers in neural architecture search also adopt RL to search neural network configurations that can give the best accuracy on validation sets \cite{zoph2016neural,pham2018efficient}, which is comparable to our search of camera configuration. 

\section{Methodology}
\label{sec:method}
\subsection{Problem Formulation}
To investigate camera configuration generation for multi-view pedestrian detection, we consider an interactive formulation. As shown in Fig.~\ref{fig:overview}, 
on the left, the \underline{\textit{agent}}, camera configuration generator $g\left(\cdot\right)$ predicts camera configuration as its action $a$. On the right, the \underline{\textit{environment}} takes as input the action (camera configuration), renders images, and feeds the images to the detection network $f\left(\cdot\right)$. Finally, it returns the detection accuracy as reward $r$ and considers the updated camera configuration as state $s$. 
It is noteworthy that, unlike most environments, the detection network can also be jointly trained, resulting in non-static rewards throughout the search process.

Mathematically, we define the configuration $\bm{c} \in \mathbb{R}^7$ for a camera as a 7-dimensional vector,
\begin{equation}
    \bm{c}=\left(x,y,z, \cos{\psi},\sin{\psi},\theta,\alpha\right), 
    \label{eq:config}
\end{equation}
where $x,y,z$ denotes the 3D location for the camera, $\psi$ and $\theta$ denote the camera yaw and pitch angle, and $\alpha$ denotes the camera FoV. 
Note that in the configuration $\bm{c}$, we use the cosine and sine values for the yaw angle $\psi\in\left[0,2\pi\right]$ to avoid having two different values ($0$ and $2\pi$) for the same angle \cite{zhou2019continuity}. 
The camera configuration $\bm{c}$ can be fed into the simulation engine to render a camera view $\bm{x}$. 

In our interactive formulation, we consider the sequential generation of configurations for $N$ cameras. At a specific time step $t\in \left\{0,\cdots,N-1\right\}$, there are $t$ cameras whose configurations are already decided. For Markovian modelling, we represent the state $s$ as, 
\begin{equation}
    s_t=
\begin{cases}
    \emptyset,& \text{if } t=0\\
    \left(\bm{c}_1,\cdots,\bm{c}_t\right),              & \text{otherwise}
\end{cases},
    \label{eq:state}
\end{equation}
where $\emptyset$ denotes an empty set. The action $a$ taken at this time step determines the configuration for the next camera,
\begin{equation}
    a_t=\bm{c}_{t+1}.
    \label{eq:action}
\end{equation}
And the reward $r$ for each time step is set as, 
\begin{equation}
    r_t=
\begin{cases}
    0,& \text{if } t< N-1\\
    \text{MODA}\left(f\left(\bm{x}_1,\cdots,\bm{x}_N\right),\bm{y}\right),              & \text{otherwise}
\end{cases},
    \label{eq:reward}
\end{equation}
where $\text{MODA}\left(\cdot, \cdot\right)$ denotes the Multi-Object Detection Accuracy \cite{kasturi2008framework}, the primary metric for multi-view pedestrian evaluation \cite{chavdarova2018wildtrack,hou2020multiview}; $f\left(\cdot\right)$ denotes the multi-view pedestrian detection network; $\bm{x}_1,\cdots,\bm{x}_N$ are the rendered camera views from $N$ cameras; and $\bm{y}$ is the corresponding ground truth 3D locations for pedestrians.

\subsection{CarlaX Environment}
\label{secsec:carlax}

In order to investigate how effective different camera configurations are quantitatively, we need a dedicated testbed. Existing datasets on multi-view detection like Wildtrack \cite{chavdarova2018wildtrack} and MultiviewX \cite{hou2020multiview} only contain static images collected from expert-designed camera rigs. To fill in the gap of controllable camera rendering and evaluation, an interactive simulation environment, CarlaX, is introduced in this work. 

\begin{figure}
\centering
\includegraphics[width=\linewidth]{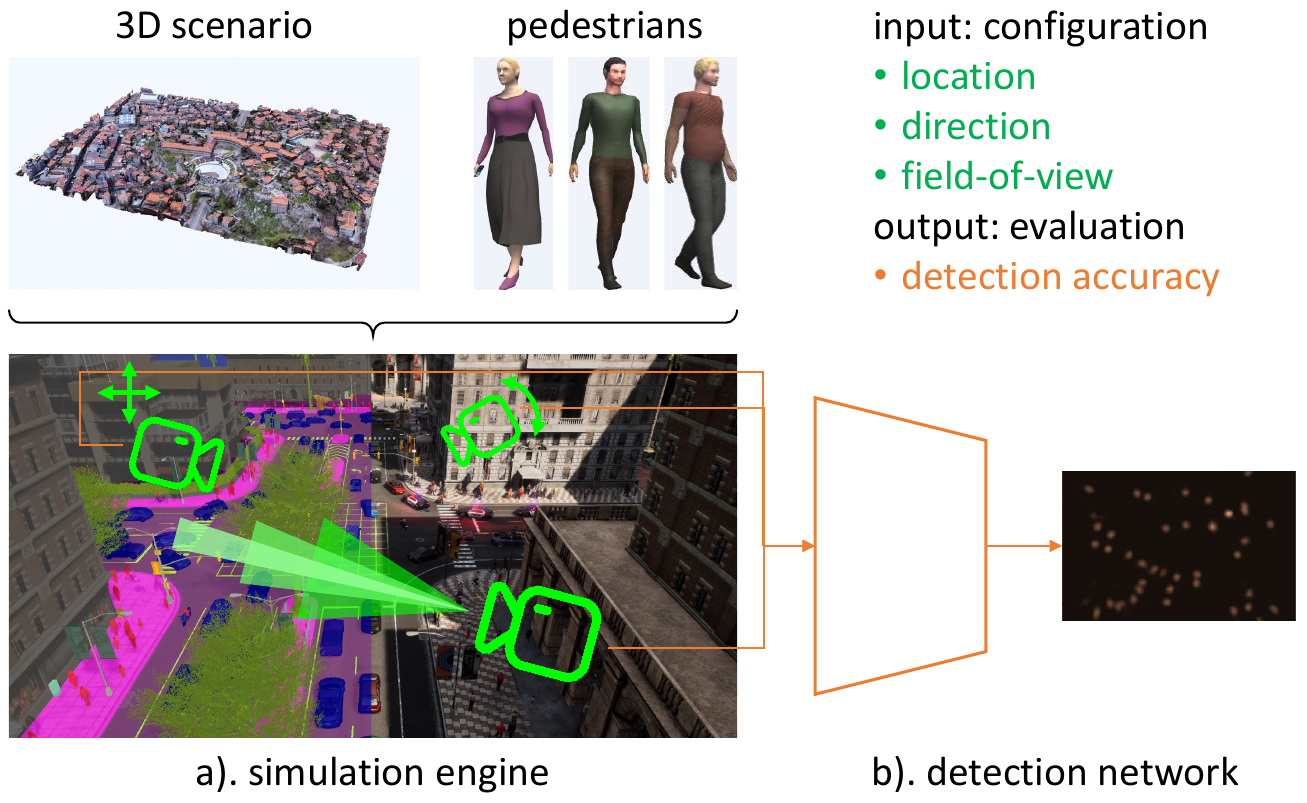}
\caption{
The interactive environment has two main components for controllable camera rendering and evaluation: a). the simulation engine and b). the multi-view pedestrian detection network. Given configurations, the simulation renders camera views, which are then fed into the detection network for quantitative evaluation.  
}
\label{fig:carlax}
\end{figure}

CarlaX connects the autonomous driving engine, Carla \cite{dosovitskiy2017carla}, with the multi-view pedestrian detection network (see Fig.~\ref{fig:carlax}). 
The \underline{\textit{simulation engine}} takes as input the camera configurations, which include the 3D location $x,y,z$, the camera yaw angle $\psi$ and pitch angle $\theta$, plus the camera FoV $\alpha$. Next, it renders the image for the 3D scenario and randomly populated pedestrians. Corresponding ground truth for 3D human locations is also preserved for further evaluation of the detection model.
The multi-view pedestrian \underline{\textit{detection network}}  $f\left(\cdot\right)$ then takes the rendered image views and estimates the 3D locations for pedestrians, which are then compared with the ground truth for evaluation. 

\begin{figure}
\centering
\includegraphics[width=\linewidth]{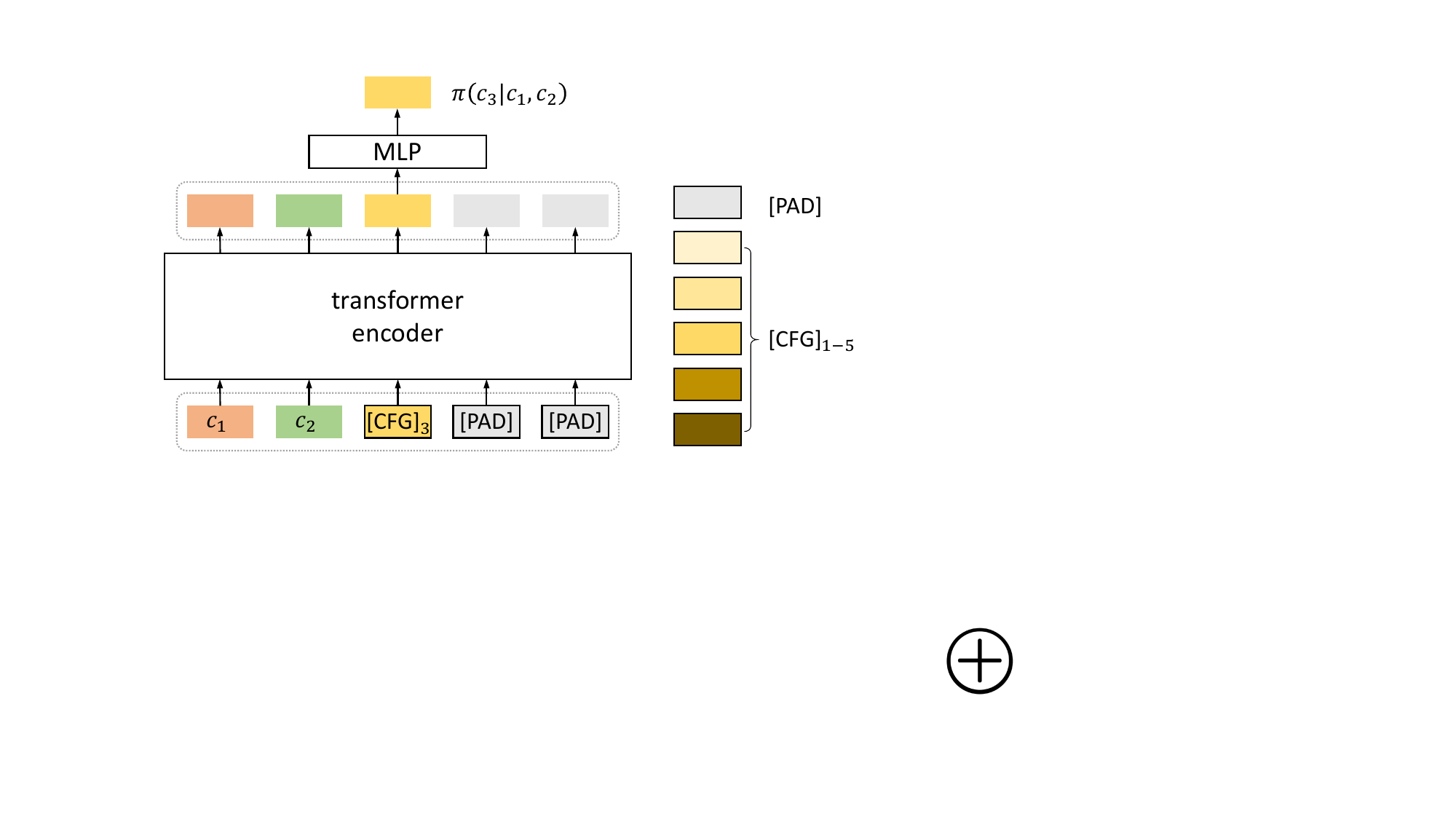}
\caption{
Architecture for camera configuration generator. For a time step $t=2$ when configuring a total $N=5$ cameras, we feed a fixed length input into the transformer encoder, consisting of configurations $\bm{c}_1$ and $\bm{c}_2$ for the first two cameras, a specific prediction token [CFG]$_3$ for the third camera, and two padding tokens [PAD]. Only the output for the prediction token [CFG]$_3$ is then fed into a multi-layer perceptron (MLP) to estimate the action for this time step $a_2=\bm{c}_{3}$, which is the configuration for the third camera. Solid black borders indicate learnable parameters. 
}
\label{fig:transformer}
\end{figure}

\subsection{Camera Configuration Generation}
\label{secsec:network}

It is our goal to learn a policy network $g\left(\cdot\right)$  that can predict camera configurations based on the sequential state representation in Eq.~\ref{eq:state}.
Importantly, the prediction for the next camera's configuration should be \textit{invariant} to the specific sequential order of the previous camera configurations, \eg, for $t=2$, $s_2 = \left(\bm{c}_1, \bm{c}_2\right)$, the policy for the next action $\pi\left(a_2|\left(\bm{c}_1, \bm{c}_2\right)\right) = \pi\left(a_2|\left(\bm{c}_2, \bm{c}_1\right)\right)$. 

To make the policy $\pi\left(a|s\right)$ \textit{invariant} to the sequence order, we adopt the transformer architecture \cite{vaswani2017attention}. Unlike fully connected layers where the concatenation order of the state tokens can affect its results, self-attention operations in transformers are \textit{permutation invariant}, making it ideal for our usage.  
The overall architecture is demonstrated in Fig.~\ref{fig:transformer}. Specifically, we only adopt the encoder part of the transformer to avoid excessive parameters and overfitting. We also maintain $N$ embedding as configuration prediction tokens $\left\{\text{[CFG]}_1,\cdots,\text{[CFG]}_N\right\}$ and a padding token [PAD].

The forward pass of the network is depicted as follows,
\begin{equation}
    \pi\left(a_t|s_t\right) = g\left({s}_t, \text{[CFG]}_{t+1}, \text{[PAD]}\right).
\end{equation}
At a specific time step $t\in \left\{0,\cdots,N-1\right\}$, given existing $t$ camera configurations in the state vector $s_t$, we append a specific prediction token [CFG]$_{t+1}$ and pad the sequence into a total of $N$ tokens. 
No positional embedding is used to maintain the permutation invariant property. 
The transformer encoder output for the prediction token [CFG]$_{t+1}$ is then fed into a multi-layer perception (MLP) to estimate the policy $\pi\left(a_t|s_t\right)$.

\begin{algorithm}[t]
\caption{Training scheme for the proposed method.}
\label{alg:training}
\begin{algorithmic}[1]
\small
\STATE \textbf{input}: camera configuration generator $g\left(\cdot\right)$, detection network $f\left(\cdot\right)$, number of cameras $N$, 
PPO buffer size $L$, maximum steps $T$.
\STATE initialize iteration counter $i=0$ and PPO buffer $B=\emptyset$;
\WHILE{$i<T$}
\STATE randomly populate pedestrians in the simulation scenario;
\STATE initialize the state $s_0=\emptyset$;
\FOR{time step $t \in \left\{0,\ldots,N-1\right\}$}
    \STATE select action $a_{t} = \bm{c}_{t+1} \sim \pi\left(a_t|s_t\right)$ using $g\left(\cdot\right)$;
    \STATE spawn the next camera and render its view $\bm{x}_{t+1}$;
    \IF{$t<N-1$}
    \STATE set $r_t=0$;
    \ELSE
    \STATE calculate detection result $\Tilde{\bm{y}}=f\left(\bm{x}_1,\cdots,\bm{x}_N\right)$ and set $r_t=\text{MODA}\left(\Tilde{\bm{y}},\bm{y}\right)$;
    \STATE calculate detection loss and optimize $f\left(\cdot\right)$;
    \ENDIF
    \STATE update the PPO buffer $B=B\cup \left\{s_t, a_t, r_t\right\}$;
    \STATE calculate the next state $s_{t+1}=s_t\cup \left\{\bm{c}_t+1\right\}$;
    \STATE update the iteration counter $i=i+1$;
\ENDFOR
\IF{$|B|>L$}
\STATE calculate PPO loss, the two regularization terms $R_\text{diverse}$ and $R_\text{focus}$, and optimize $g\left(\cdot\right)$;
\STATE set  the PPO buffer $B=\emptyset$;
\ENDIF
\ENDWHILE
\STATE \textbf{return}: configuration generator $g\left(\cdot\right)$ and detector $f\left(\cdot\right)$.
\end{algorithmic}
\end{algorithm}

\subsection{Training Scheme}
\label{secsec:training}

In order to effectively navigate the vast action space while dealing with non-differentiable procedures in the interactive environment, we use reinforcement learning to guide the camera configuration policy. Specifically, we adopt the PPO algorithm \cite{schulman2017proximal}, a policy-gradient-based RL method for its simplicity and stability. After collecting a buffer of state $s$,  action $a$, and reward $r$, PPO estimates the specific advantage \cite{schulman2015high} for taking an action, and updates the policy network $g\left(\cdot\right)$ using its policy loss. 
During the camera search process, the detection network can also be jointly trained using the ground truth pedestrian locations. 
We show a step-by-step training scheme in Algorithm \ref{alg:training}.

Despite the effectiveness of the PPO algorithm, the policy gradient remains a little indirect compared to differentiable feedback. This can be potentially problematic considering the vast action space -- for each camera, the action space spans over $>$1000m$^3$ for camera location, $2\pi$ for the camera yaw angle, plus the pitch angle and camera FoV. In this case, we introduce two additional heuristic-based regularization terms for the camera configurations, both of which can provide direct and differentiable supervision. 

Our first heuristic is that cameras should ideally be diverse in their viewing angles. 
Mathematically, we compute the diversity regularization term as,
\begin{equation}\small
\begin{aligned}
    R_\text{diverse} = -
&\sum_{\tau}{\left\|\left(\Bar{x}-\Bar{x}_\tau, \Bar{y}-\Bar{y}_\tau\right)\right\|_2}
    \\
-&\sum_{\tau}{\left\|\left(\cos{\psi}-\cos{\psi_\tau}, \sin{\psi}-\sin{\psi_\tau}\right)\right\|_2},
\end{aligned}
    \label{eq:diverse}
\end{equation}
where $\tau$ denotes a total of $t$ previous cameras. By minimizing this regularization term, cameras are encouraged to provide different viewing angles. 

Our second heuristic is that cameras should focus on the scenario and avoid looking outside of the scenario. For the regularization term, we consider the following,
\begin{equation}
    R_\text{focus} = \frac{\left\|\left(\Bar{x}+\delta\cos{\psi}, \Bar{y}+\delta\sin{\psi}\right)\right\|_2 - \left\|\left(\Bar{x}, \Bar{y}\right)\right\|_2}{\delta},
    \label{eq:focus}
\end{equation}
where $\left\|\cdot\right\|_2$ denotes the Euclidean norm; $\Bar{x}, \Bar{y} \in \left[-1,1\right]$ are normalized camera $xy$ coordinates; $\delta$ is a small hyper-parameter. Promoting a smaller regularization term $R_\text{focus}$ can penalize camera yaw angle $\psi$ for looking outside of the scenario when the positioning is already offset to that side.

\begin{table}[t]
\caption{Datasets comparison for multi-view pedestrian detection.}
\label{tab:dataset}
\resizebox{\linewidth}{!}{
\begin{tabular}{l|c|c|c|c}
\toprule
           & \#camera  & frames & area        & \#person \\ \hline
Wildtrack & 7       & 400    & $12\times 36$ m$^2$      & 20       \\ \hline
MultiviewX & 6       &  400    & $16\times 25$ m$^2$       & 40       \\ \hline
CarlaX     & up to 6 &  400    & up to $45\times 39$ m$^2$ & up to 80 \\ \bottomrule
\end{tabular}
}
\end{table}

\section{Experiment}
\label{sec:experiment}


\subsection{Experimental Settings}
\textbf{Camera parameters and action space.} In our study, we consider basic RGB cameras that can be mounted at a specific location and direction with a fixed focal length. In terms of the action space for camera configurations (Eq.~\ref{eq:config}), the range for location $x,y,z$ is determined by specific scenarios; the yaw angle $\psi\in\left[0,360\degree \right]$ spans the entire unit circle; the pitch angle $\theta\in\left[-30\degree ,30\degree \right]$ allows the camera to adjust for different mounting height; and the FoV $\alpha\in\left[30\degree,120\degree \right]$ covers the common focal lengths \cite{vigderman_2023_where}.

\begin{table*}[]
\caption{Detection performance comparison on seven CarlaX scenarios. We compare the proposed method with three human experts, random search, and maximum FoV coverage, under various numbers of cameras and location Degrees of Freedom (DoFs). The results of three optimization-based methods are the average of 5 independent runs with standard deviation. MODA (\%) is reported. The best results are shown in \textbf{bold}. 
Our proposed approach is better than the best human configuration.
}
\label{tab:performance}
\vspace{-2mm}
\centering
\resizebox{\linewidth}{!}{
\begin{tabular}{l|c|c|ccc|cc|c}
\toprule
\multirow{2}{*}{Scenario}  & \multirow{2}{*}{\#camera} & \multirow{2}{*}{Location DoF} & \multicolumn{6}{c}{Camera configuration designs}                                \\ \cline{4-9}
         &  &  & Expert 1 & Expert 2 & Expert 3 & random search & max coverage & proposed \\ \hline
\texttt{Town03cafe}       & 4 & $42\times 10\times 6$ m$^3$  & 71.3     & 85.8     & 77.4     & 25.9 $\pm$ 10.2              & 60.7 $\pm$ 3.0            & \textbf{91.7} $\pm$ 0.7   \\
\texttt{Town03park}       & 3 & $45\times 39\times 7$ m$^3$  & 69.9     & 82.6     & 70.0     & 57.4 $\pm$ 5.0              & 65.4 $\pm$ 2.7             & \textbf{92.0} $\pm$ 0.5     \\
\texttt{Town04building}   & 4 & $15\times 30\times 8$ m$^3$  & 72.4     & 89.0     & 82.9     & 46.3 $\pm$ 7.0              & 52.1 $\pm$ 4.4              & \textbf{95.8} $\pm$ 0.4     \\
\texttt{Town04crossroad}  & 4 & $20\times 20\times 4$ m$^3$  & 95.3     & 94.2     & 95.0     & 90.2 $\pm$ 4.4              & 74.4 $\pm$ 2.2             & \textbf{98.6} $\pm$ 0.2     \\
\texttt{Town05building}   & 4 & $15\times 25\times 8$ m$^3$  & 91.2     & 94.4     & 72.9     & 66.5 $\pm$ 6.3              & 76.5 $\pm$ 1.9             & \textbf{98.5} $\pm$ 0.2     \\
\texttt{Town05market}     & 6 & $56\times 22\times 4$ m$^3$  & 48.1     & 58.6     & 44.3     & 21.3 $\pm$ 4.2              & 15.3 $\pm$ 3.9             & \textbf{68.7} $\pm$ 1.5     \\
\texttt{Town05skyscraper} & 3 & $28\times 11\times 6$ m$^3$  & 92.9     & 90.4     & 85.2     & 72.9 $\pm$ 6.9              & 79.0 $\pm$ 1.9              & \textbf{97.9} $\pm$ 0.1     \\ \bottomrule
\end{tabular}
}
\end{table*}

\textbf{Simulation scenarios.} We introduce seven distinct scenarios in our interactive environment, CarlaX. 
An overall statistical comparison between CarlaX and existing datasets Wildtrack \cite{chavdarova2018wildtrack} and MultiviewX \cite{hou2020multiview} can be found in Table~\ref{tab:dataset}. 
By default, in each scene, we randomly populate up to 40 walking people and up to 10 groups of 2 to 4 chatting persons. Although CarlaX can generate infinite pedestrian distributions, we limit them to 400 distinct patterns (frames)  to mimic existing datasets. 
\begin{itemize}
    \item \texttt{Town03cafe}: a small coffee shop with tables, located in the middle of four apartment buildings. 
    \item \texttt{Town03park}: a park with trees, slopes, swings, and a lot of open space, making it difficult to cover. 
    \item \texttt{Town04building}: a plaza between buildings with occlusions from a pergola (sitting area with pillars supporting an open lattice top) and sunshades of street shops. 
    \item \texttt{Town04crossroad}: a crossroad with traffic lights. It is very open and easy for pedestrian detection.
    \item \texttt{Town05building}: a plaza between buildings but without severe occlusions. 
    \item \texttt{Town05market}: the most challenging scenario, covering a flea market with multiple rows of shops and food stands. 
    \item \texttt{Town05skyscraper}: an area between high buildings with a giant tree, limiting visibility from certain angles.
\end{itemize}
As shown in Table~\ref{tab:performance}, each scene has a different Degree of Freedom (DoF) for the camera location ($x,y,z$ in camera configuration $\bm{c}$) and pedestrian spawning (only two out of three dimensions). Specifically, due to the nature of the scenario, we set different ranges for the camera mounting height ($z$), since tall buildings can naturally support higher mounting points while newspaper stands and flea markets cannot support very high mounting points.

\textbf{Baselines} considered in this study are listed as follows. 
\begin{itemize}
    \item \underline{\textit{Human expert}}: we recruit three human experts to position the cameras for each scenario. 
    \item \underline{\textit{Random search}}: a similar computation budget is given to randomly generate camera configurations, and \textit{greedily} chooses the configuration that gives the best detection accuracy on one frame of training images. 
    \item \underline{\textit{Maximum FoV coverage}}: following previous works \cite{erdem2006automated,sun2021learning}, we consider the FoV coverage of ground plane building floor plans as a heuristic for camera placement study. During training, we follow previous work and do not spawn pedestrians into the scenario, so as to focus on the occlusion introduced by the static objects like tables, chairs, trees, buildings, \etc. The configuration with the highest FoV coverage for the ground plane is selected. 
\end{itemize}
We do not consider view selection methods for 3D reconstruction \cite{fan2016automated,zhou2020offsite,lee2022uncertainty}, since they serve a very different purpose (recognition of foreground objects v.s. reconstruction of static background) and often require hundreds of views. 
%

\textbf{Multi-view pedestrian detection network.}
We choose MVDet \cite{hou2020multiview} as the detector network $f\left(\cdot\right)$ in this study, but it can be changed into more advanced architectures like SHOT \cite{song2021stacked} or MVDeTr \cite{hou2021multiview}. 

\textbf{Evaluation metric.} Evaluation metrics for multi-view pedestrian detection include multi-object detection accuracy (MODA, the primary indicator), multi-object detection precision (MODP, evaluation for localization error), and precision and recall (combined they give MODA) \cite{kasturi2008framework}. Following previous work \cite{hou2020multiview,hou2021multiview,song2021stacked}, we consider MODA as the primary indicator for performance, since detecting the targets (MODA, precision, recall) is more important than having the most accurate localization (MODP), and MODA accounts for both precision and recall.

\subsection{Implementation Details}
For our camera configuration generator, we use three layers of transformer encoder with a feature dimension of 128.

We train all networks for 50 epochs (which roughly gives $T=50,000$ training steps when there are three cameras in the scene) using the Adam optimizer \cite{kingma2015adam}. We use learning rates of $1\times10^{-5}$ and $1\times10^{-4}$ for $f\left(\cdot\right)$ and $g\left(\cdot\right)$, respectively. For the PPO training, we use a buffer size of $L=1024$ and a mini-batch size of 128, and train on each recorded experience 10 times. 
Regarding hyperparameters, we set the weight for the two regularization terms as $0.1$. 

All experiments are run on one RTX-3090 GPU and averaged across 5 repetitive runs, taking roughly 8 hours each.

\subsection{Experimental Results}

\begin{figure*}
\centering
    \begin{subfigure}[b]{\linewidth}
        \includegraphics[height=0.087\linewidth]{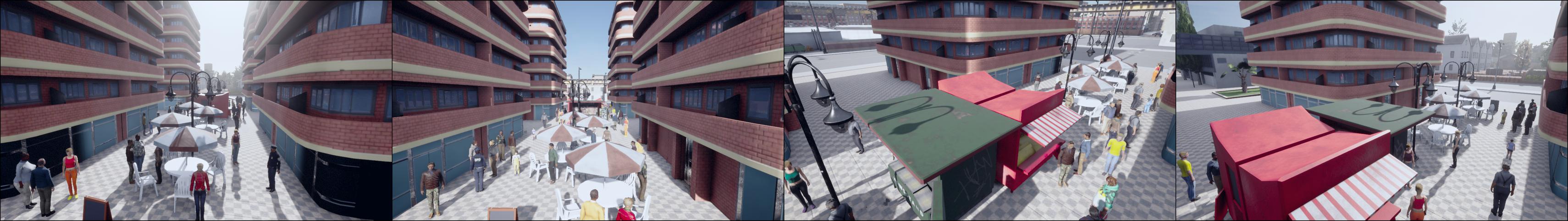} \hfill
        \includegraphics[height=0.087\linewidth]{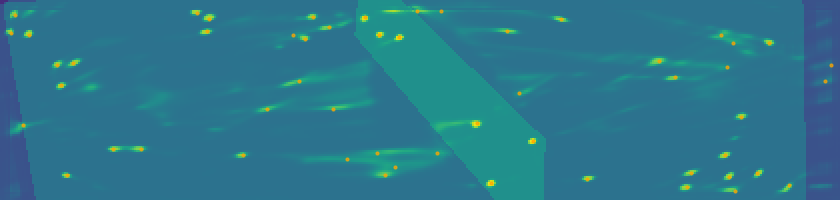} 
        
        \includegraphics[height=0.087\linewidth]{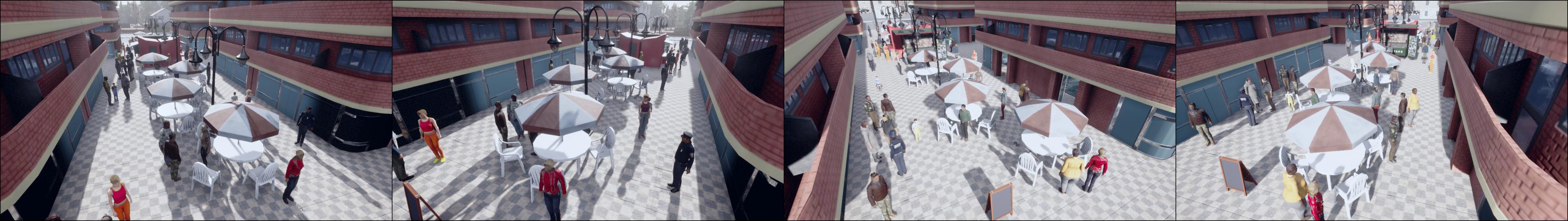} \hfill
        \includegraphics[height=0.087\linewidth]{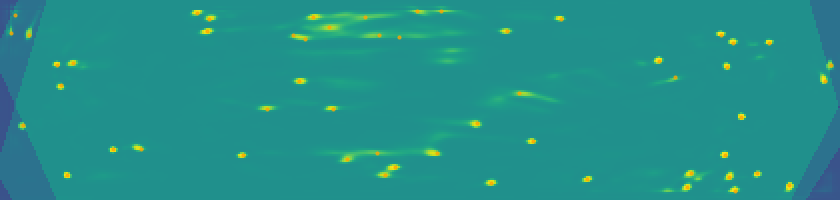} 
        
        \includegraphics[height=0.087\linewidth]{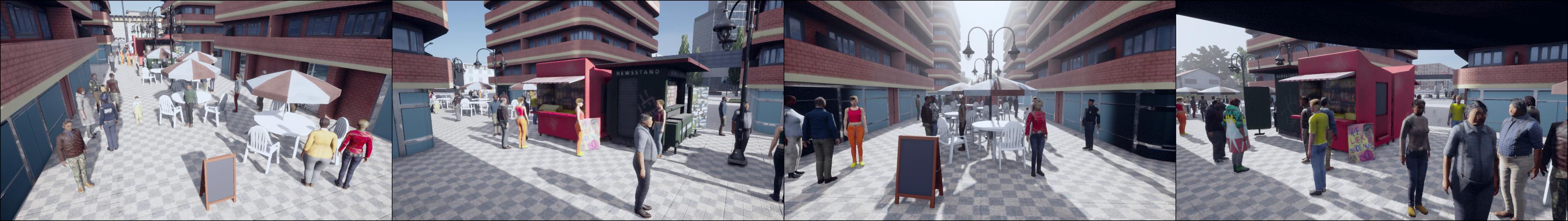} \hfill
        \includegraphics[height=0.087\linewidth]{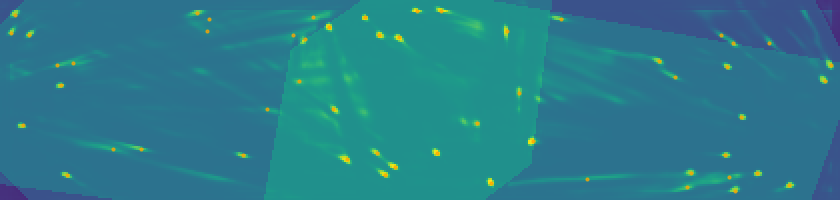} 
        
        \includegraphics[height=0.087\linewidth]{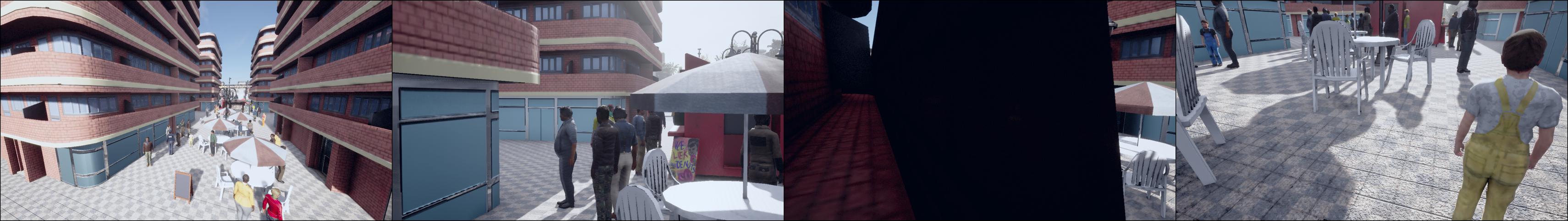} \hfill
        \includegraphics[height=0.087\linewidth]{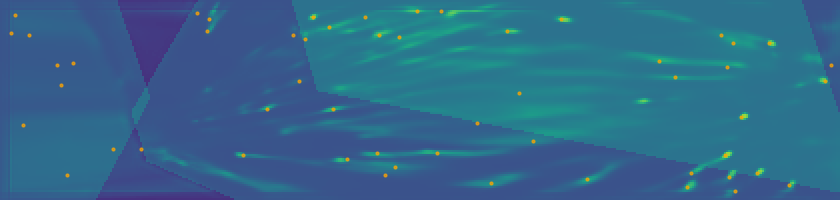} 
        
        \includegraphics[height=0.087\linewidth]{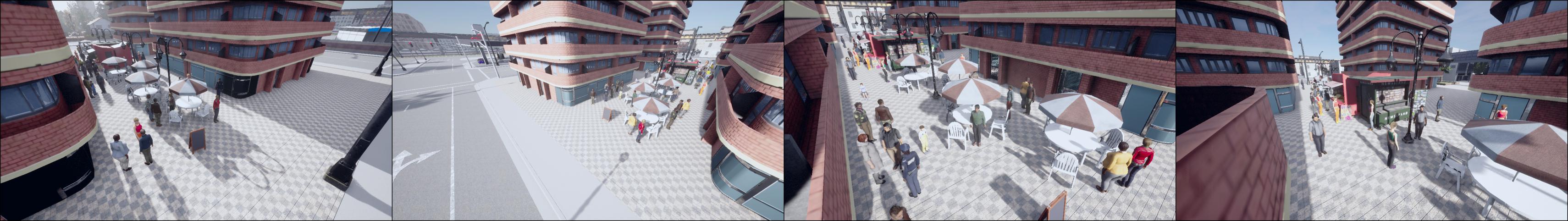} \hfill
        \includegraphics[height=0.087\linewidth]{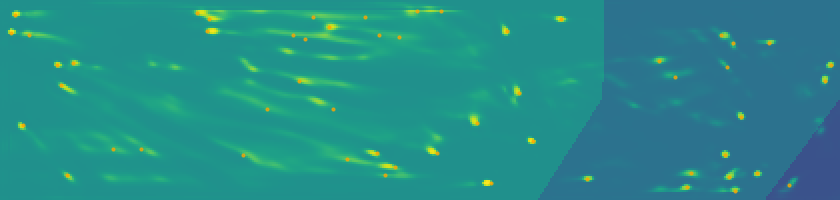} 
        
        \includegraphics[height=0.087\linewidth]{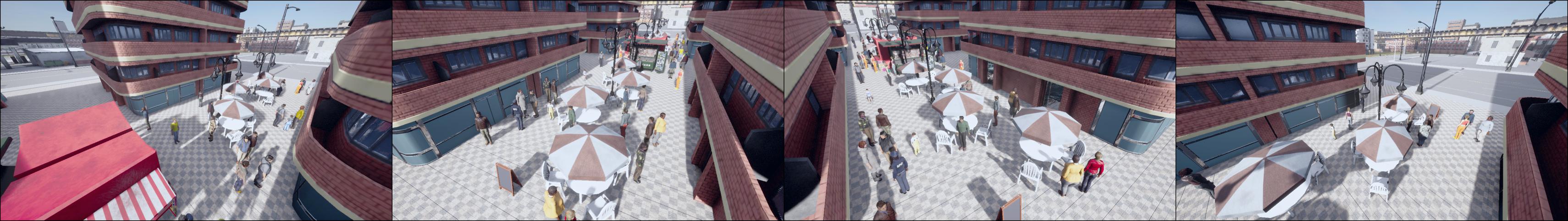} \hfill
        \includegraphics[height=0.087\linewidth]{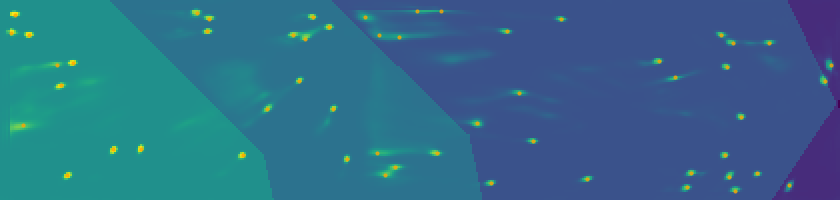} 
    \end{subfigure}
\vspace{-1mm}
\caption{
Comparison between camera configurations on \texttt{Town03cafe}. From top to bottom, we have camera views from Expert 1, Expert 2, Expert 3, random search, max coverage, and the proposed method. We also show FoV coverages and detection results (heatmap overlay) for pedestrians (orange dots). For more visualizations of other scenarios, please see the Appendix.  
}
\label{fig:comparison}
    \vspace{-2mm}
\end{figure*}

We compare the multi-view pedestrian detection performance in Table~\ref{tab:performance} and Fig.~\ref{fig:comparison}. Overall, we find the proposed method consistently outperforms the human experts and baseline methods. We have the following findings. 

First, we find that different \underline{\textit{human experts}} achieve quite diverse performance (Table~\ref{tab:performance}) due to their distinct preferences (Fig.~\ref{fig:comparison}): Expert 1 tends to position cameras along the sides; Expert 2 likes to position cameras at the corners of the scenario; Expert 3 combines close-up views with wide-angle views, possibly reflecting some conscious or unconscious bias or preferences. Arguably, these expert configurations all make sense, but which of them actually yields the best results requires experimental verification, something often overlooked in human expert designs \cite{zong2018method,hori1997traffic,vigderman_2023_where,conceptdraw_camera}. This also verifies the need for an interactive playground to import 3D scene scanning and quantitatively test out the detection accuracy of the potential configurations, something we provide in our CarlaX environment.

Second, we find that \underline{\textit{random search}} cannot efficiently locate the helpful views and is very noisy. In fact, with multiple experiments showing test accuracy first increasing and then decreasing, we find the detection accuracy on {one frame} of training images is not a very robust indicator, as the test set might display rather different pedestrian distributions. Also, the large variance between different repeats of random search (Table~\ref{tab:performance}) indicates such simple approach is not suitable for vast action spaces. 

\begin{figure}
\centering
    \begin{subfigure}[b]{0.48\linewidth}
        \includegraphics[width=\linewidth]{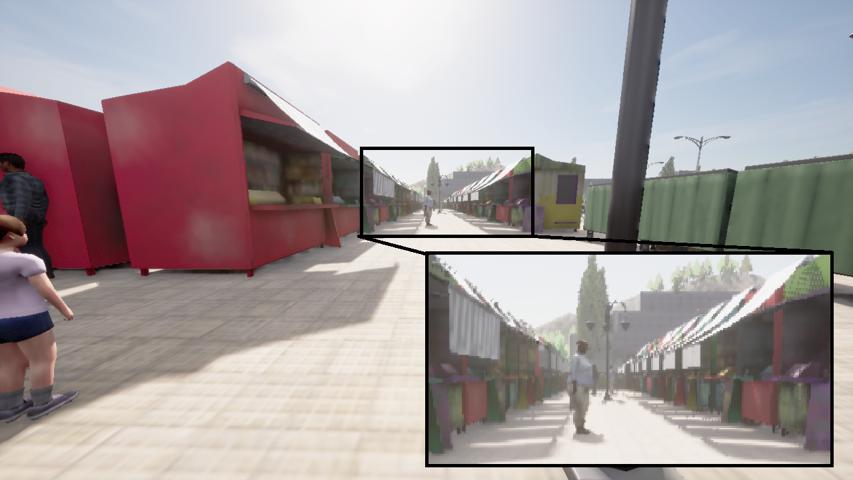} 
        \includegraphics[width=\linewidth]{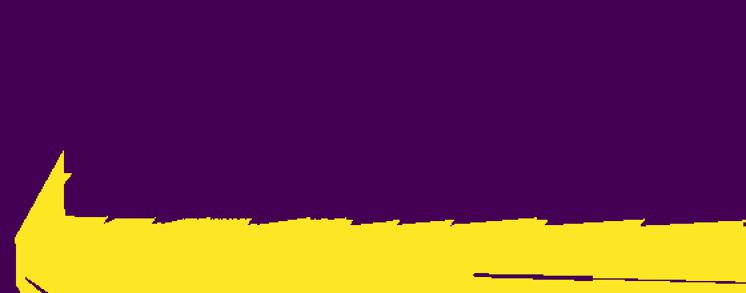}
        \caption{Max FoV coverage}
    \end{subfigure}    
    ~
    \begin{subfigure}[b]{0.48\linewidth}
        \includegraphics[width=\linewidth]{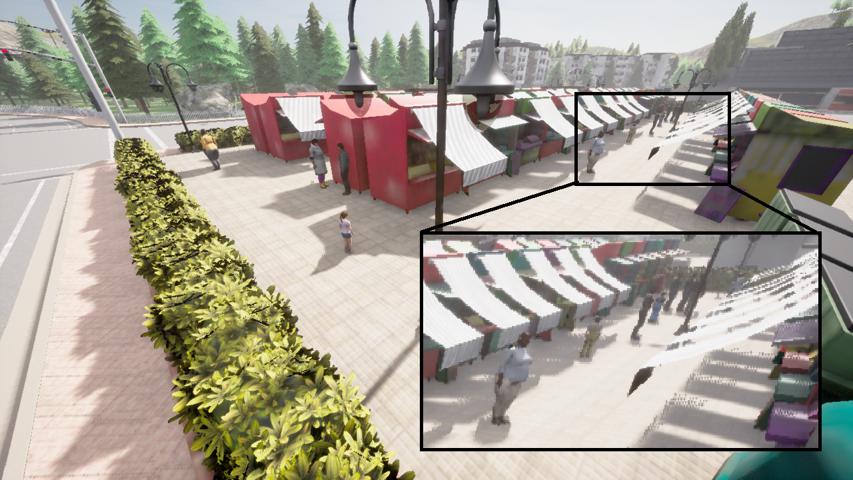} 
        \includegraphics[width=\linewidth]{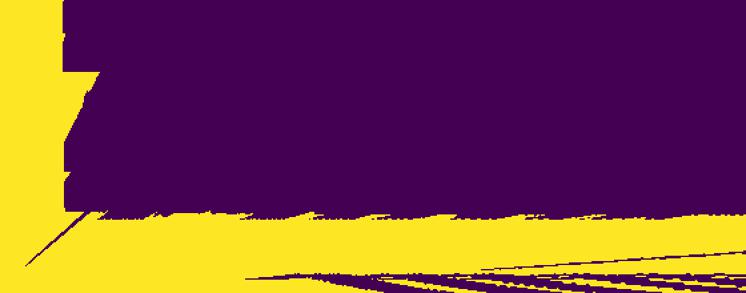}
        \caption{Proposed method}
    \end{subfigure}
    \vspace{-2mm}
\caption{
Comparison on \texttt{Town05market}. Although (a) provides good coverage (bottom) of the aisle between shops, it struggles with occlusions introduced by pedestrians (top). 
}
\label{fig:visibility}
    \vspace{-2mm}
\end{figure}

\begin{figure*}
\vspace{-3mm}
\begin{minipage}[t]{0.66\textwidth}
\includegraphics[width=0.48\linewidth]{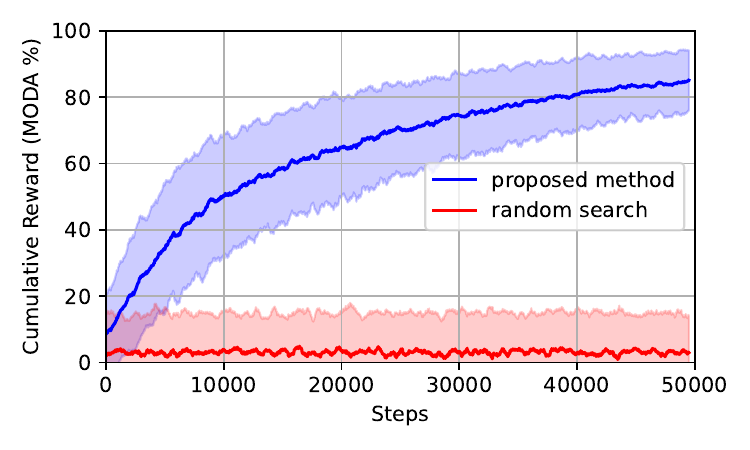}
\;
\includegraphics[width=0.48\linewidth]{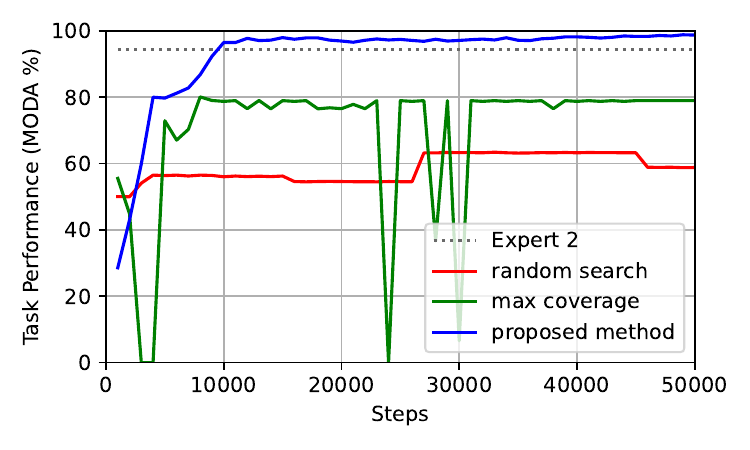}
\vspace{-5mm}
\captionof{figure}{Convergence on \texttt{Town05building}. On the left, we show the cumulative reward of the training set MODA (max coverage \cite{erdem2006automated,sun2021learning} uses floor plan coverage as reward, not MODA). On the right, we show the test set MODA for searched configurations.}
\label{fig:converge}
\end{minipage}
\;
\begin{minipage}[t]{0.32\textwidth}
\vspace{-32mm}
\captionof{table}{Variant study. The MLP generator fails to converge on \texttt{Town05building}, denoted as Did Not Converge (DNC).}
\label{tab:variant}
\centering
\small
\setlength{\tabcolsep}{4pt}
\begin{tabular}{l|cc}
\toprule
Variants                & T03park & T05bldg \\ \hline
proposed method & 92.0           & 98.5              \\ \hline
fixed detector   & 83.9            & 97.1               \\ 
MLP generator   & 36.8           & DNC               \\ 
w/o $R_\text{diverse}$  & 90.0           & 97.9               \\
w/o $R_\text{focus}$    & 88.2           & 98.1               \\
\bottomrule
\end{tabular}
\end{minipage}
\end{figure*}

\begin{figure}
\centering
\includegraphics[width=0.48\linewidth]{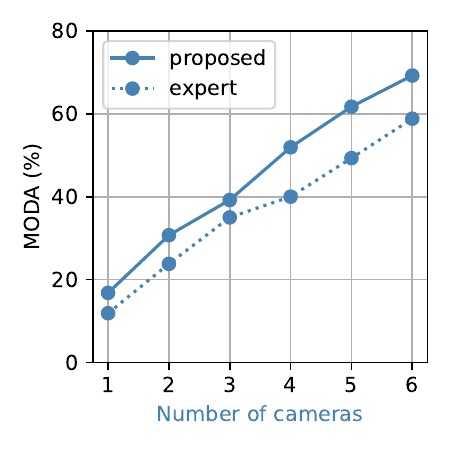}
\includegraphics[width=0.48\linewidth]{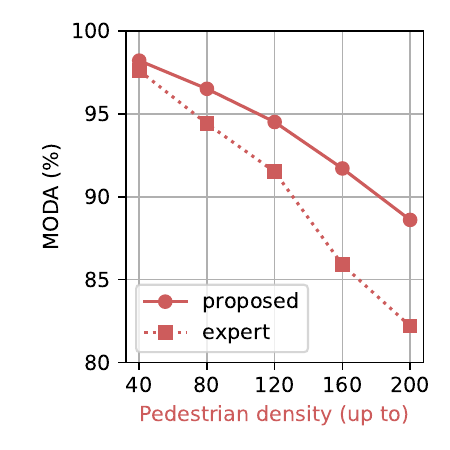}
\vspace{-3mm}
\captionof{figure}{Performance scaling over different numbers of cameras (left, \texttt{Town05market}) and different pedestrian density (right, \texttt{Town05building}).}
\label{fig:curves}
\vspace{-3mm}
\vspace{-3mm}
\end{figure}

Third, we find that \underline{\textit{maximum coverage}} is a relatively good heuristic, but cannot consistently outperform human experts in terms of task performance (Table~\ref{tab:performance}) when under vast action spaces. We demonstrate two views in Fig.~\ref{fig:visibility}, both of which provide similar FoV coverage when the scenario is empty. However, when the scene is populated, they offer different levels of help for detection under occlusion. Since FoV coverage maximization methods \cite{erdem2006automated,sun2021learning} run the optimization in empty scenes (populating the scene will introduce random occlusions, which poses additional challenges for the optimization process), it is difficult for them to guarantee good detection accuracy under occlusion. 

Last but not least, the proposed optimization can steadily converge \ref{fig:converge} and demonstrates consistent performance across multiple scenarios. It learns complex strategies, including maximizing FoV coverage, selecting diverse viewpoints, minimizing the occlusion from both static scenes (\eg, the sunshade in Fig.~\ref{fig:comparison}) and other pedestrians, and collaborating with other cameras. Compared to human experts, it also shows to be the most advantages in the most challenging scenario with most cameras, \ie, +10.1\% MODA in \texttt{Town05market}.




\subsection{Discussions}

\textbf{Fixed detection network.} As shown in Table.~\ref{tab:variant}, fixing the detector $f\left(\cdot\right)$ limits the overall system performance. However, under this setting, the RL search alone still achieves competitive performance to the best-performing human expert. Such results verify the effectiveness of both the RL search method itself, and the necessity of jointly training the detection network. 

\textbf{Choice of generator architecture.}
We find that using MLP for the camera configuration generator network $g\left(\cdot\right)$ results in worse performance (Table~\ref{tab:variant}) due to the dependency on the configuration sequence order: during RL training, the agent might sample similar configurations in different orders, which are the same to the simulation rendering. Directly concatenating the state vector $s$ can result in two very different feature vectors $\bm{c}_1\mdoubleplus \bm{c}_2$ and  $\bm{c}_2\mdoubleplus \bm{c}_1$, where $\mdoubleplus$ denotes the concatenation operation, causing additional difficulties for training. 
The proposed transformer architecture (Section~\ref{secsec:network}), on the other hand, is \textit{permutation invariant} and thus easier to train. 

\textbf{Influence of the regularization terms.}
In Table~\ref{tab:variant}, we find that removing either of the regularization terms leads to performance drops. In fact, compared to running the official PPO implementation \cite{stable-baselines3} on classic RL environments like Atari \cite{mnih2013playing} and Mujoco \cite{todorov2012mujoco} for 2M steps, running the RL algorithm over 50k training steps is rather short. However, the complex simulation update in CarlaX makes more training steps very costly, and the differentiable approximation of policy gradient \cite{sutton2018reinforcement} is not the most efficient signal. Thus, having differentiable regularization terms (Section \ref{secsec:training}) that can filter out unlikely choices turns out to be quite handy. 

\textbf{Performance scaling.}
In Fig.~\ref{fig:curves}, we show the system performance over different numbers of cameras and pedestrian density and compare them with the best human expert. 

For the number of cameras, we find that using fewer cameras can be even more challenging for this scene since they cannot cover the entire area and have no other camera to collaborate with when dealing with occlusions. Increasing the number of cameras alleviates this issue, and the search method maintains its lead over the best human expert across different numbers of cameras. 

For the pedestrian density, we find that not only does the proposed layout outperform the best human expert, but its advantages grow even larger in more crowded scenes. 

Additionally, from Table~\ref{tab:performance}, we also find the proposed method to have bigger advantages over human experts in scenes that have more cameras (\eg, \texttt{Town05market}), more area (\eg, \texttt{Town03park}), or more occlusions (\eg, \texttt{Town03cafe}, \texttt{Town04building}, \texttt{Town05market}). 

\textbf{Generalization to the real world.} 
For real-world applications, we can first scan the scenario into 3D with smartphones \cite{iphone14pro,davison2007monoslam,mur2015orb}; and then import the 3D scenario into CarlaX; and then jointly optimize both camera configurations and detection networks; and finally run domain adaptation algorithms \cite{inoue2018cross,khodabandeh2019robust} for synthetic-to-real adaptation.

\textbf{Limitation and ethics.}
In this work, we focus on multi-view pedestrian detection. Although this task touches multiple fronts including combating occlusions, reducing ambiguities, and increasing FoV coverage, it only focuses on pedestrians. In the future, we aim to extend this work to more classes like vehicles and to other tasks including 3D reconstruction. 
Moreover, we acknowledge that technologies like pedestrian detection can potentially have negative social impacts on privacy. To best avoid this issue, we conduct experiments on synthetic data. Plus, the searched camera configurations are usually from higher viewing angles (see Fig.~\ref{fig:comparison} and Fig.~\ref{fig:visibility}), further alleviating privacy issues.

\section{Conclusion}

In this paper, we extend the optimization from neural network parameters to camera configurations, and study this problem under the challenging task of multi-view pedestrian detection. Unlike previous camera placement designs that either rely on human experts or heuristic-based surrogates, we directly optimize for the task performance. To this end, we introduced an interactive environment, CarlaX, which not only enables this work but also allows for future study on camera configurations. We also designed a transformer-based configuration generator and a reinforcement-learning-based training scheme, the combination of which helps design camera configurations that can give very competitive detection accuracy. Inspired by the success of this camera configuration study, we would like to extend the optimization to other fields, including 3D reconstructions from hundreds of views and robotic vision where both camera angles and robot movements can be controlled.



{
    \small
    \bibliographystyle{ieeenat_fullname}
    \bibliography{main}
}


\newpage
\section*{Appendix}



\begin{figure*}[!htb]
\centering
        \includegraphics[height=0.151\linewidth]{figs/town03park-E0.jpg} \hfill
        \includegraphics[height=0.151\linewidth]{figs/town03park-cover-E0.png} 
        
        \includegraphics[height=0.151\linewidth]{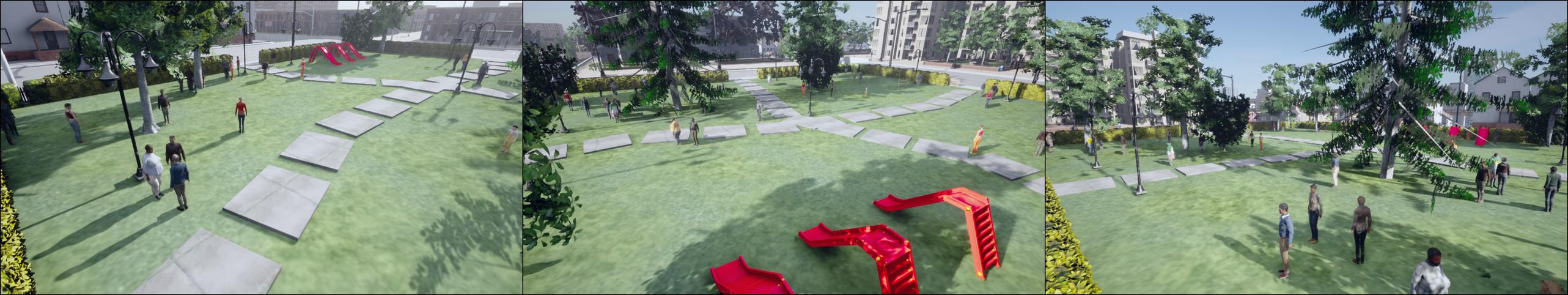} \hfill
        \includegraphics[height=0.151\linewidth]{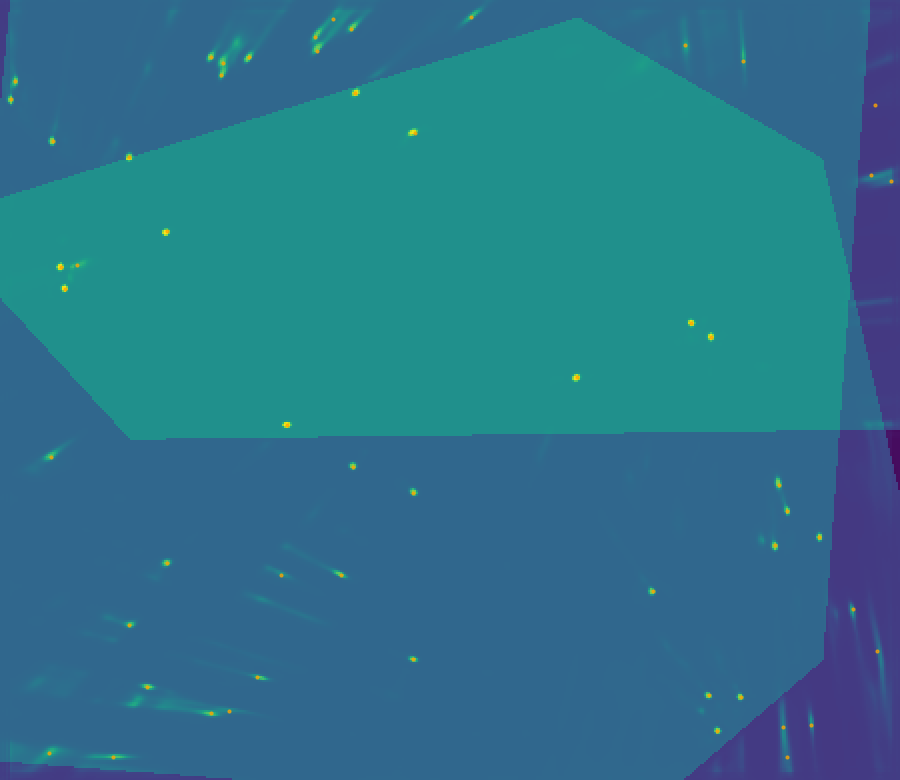} 
        
        \includegraphics[height=0.151\linewidth]{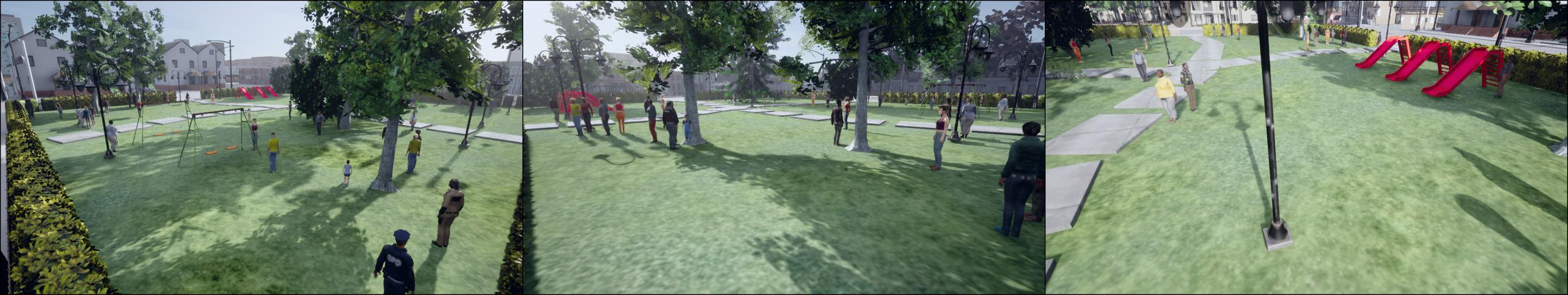} \hfill
        \includegraphics[height=0.151\linewidth]{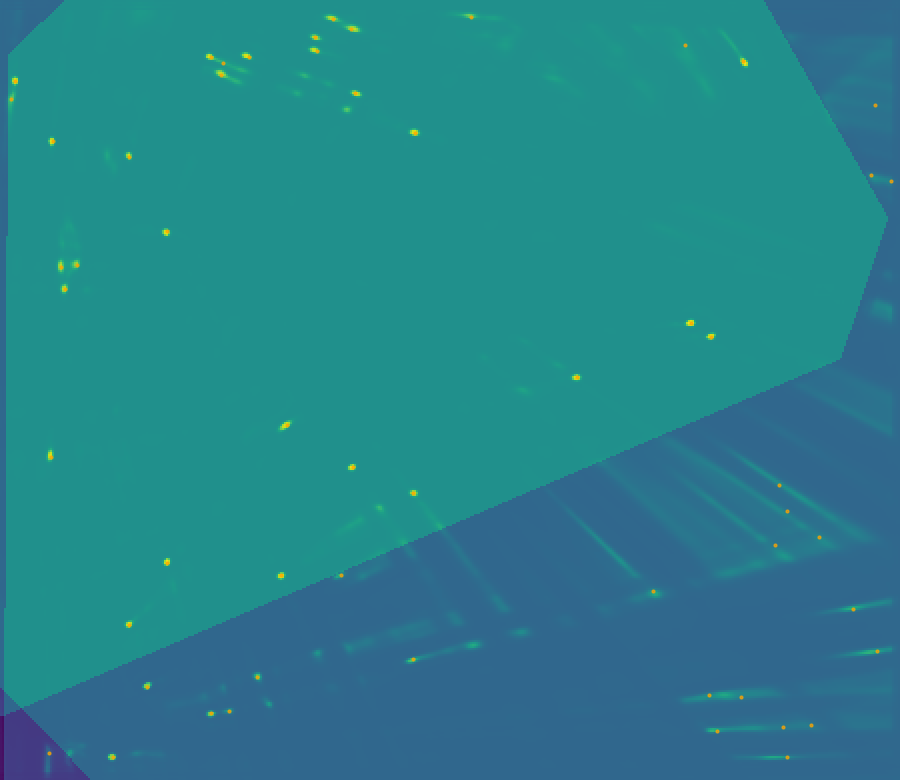} 
        
        \includegraphics[height=0.151\linewidth]{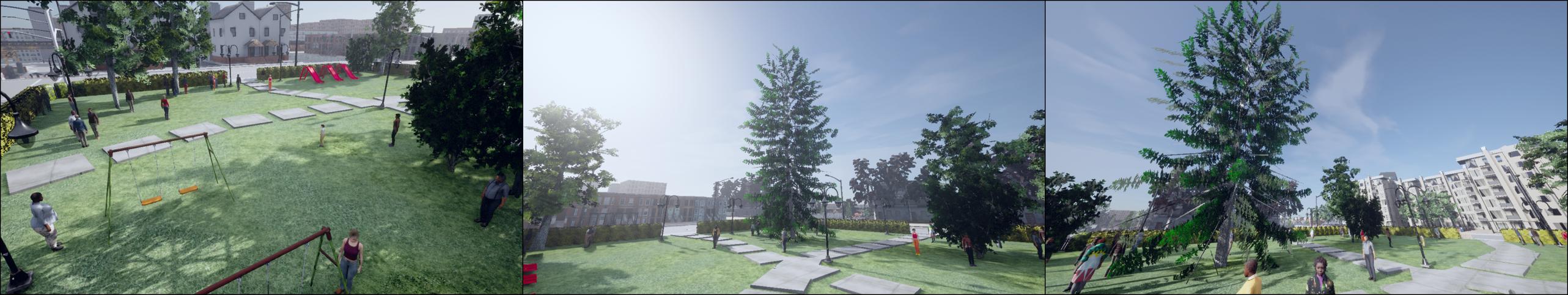} \hfill
        \includegraphics[height=0.151\linewidth]{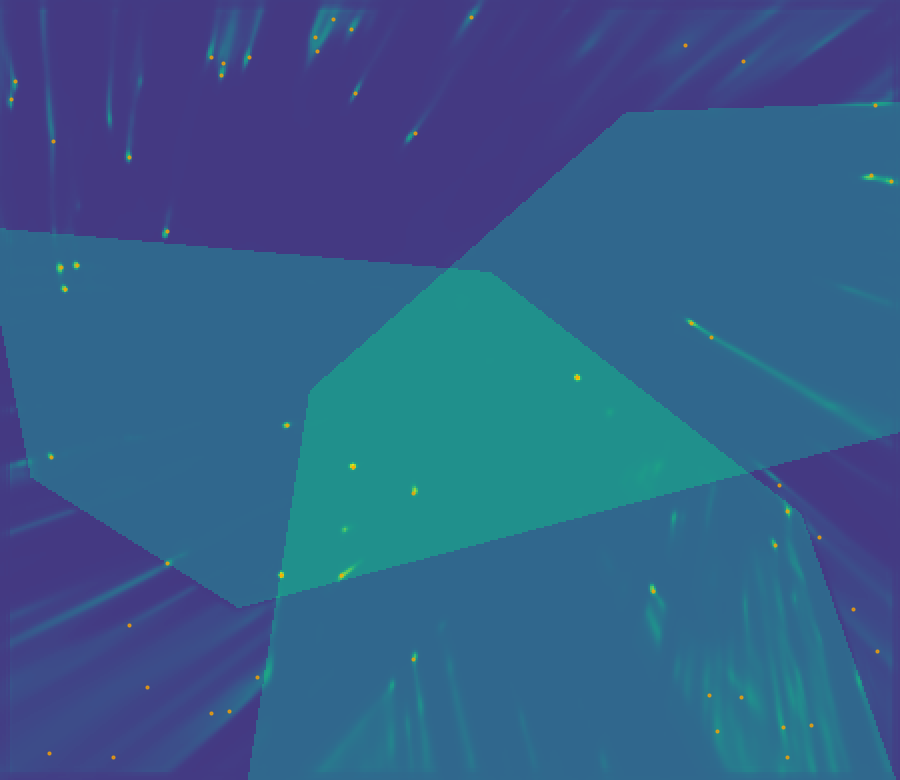} 
        
        \includegraphics[height=0.151\linewidth]{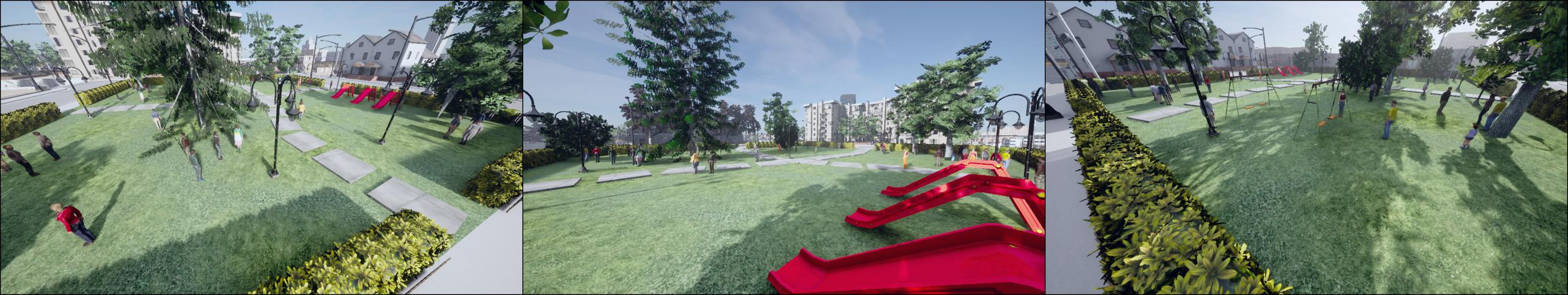} \hfill
        \includegraphics[height=0.151\linewidth]{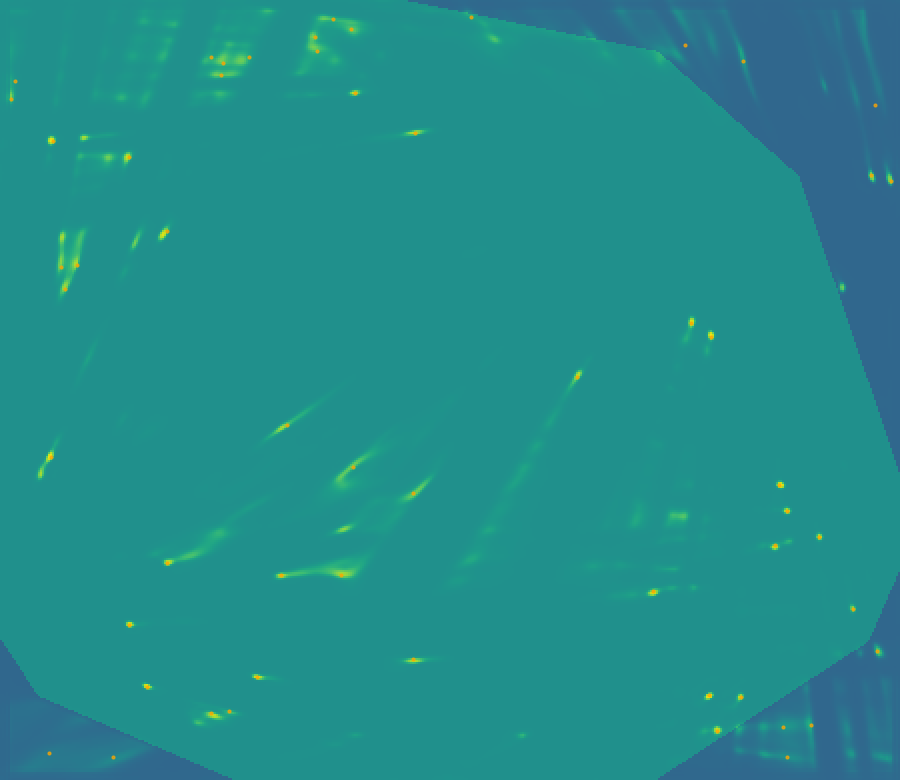} 
        
        \includegraphics[height=0.151\linewidth]{figs/town03park-J.jpg} \hfill
        \includegraphics[height=0.151\linewidth]{figs/town03park-cover-J.png} 
    
\caption{
Comparison between camera configurations on \texttt{Town03park}. From top to bottom, we have camera views from Expert 1, Expert 2, Expert 3, random search, max coverage, and the proposed method. We also show FoV coverages and detection results (heatmap overlay) for pedestrians (orange dots). 
}
\end{figure*}


\begin{figure*}
        \includegraphics[height=0.107\linewidth]{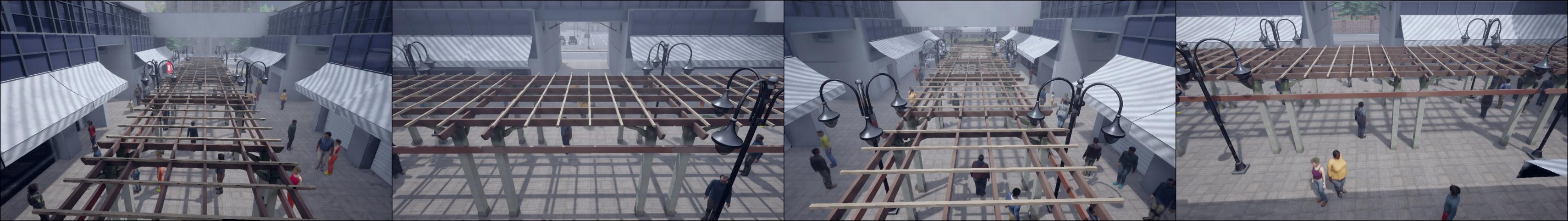} \hfill
        \includegraphics[height=0.107\linewidth]{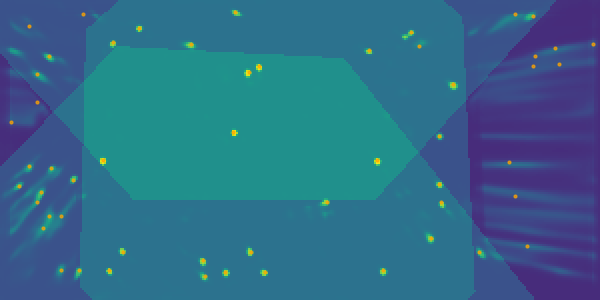} 
        
        \includegraphics[height=0.107\linewidth]{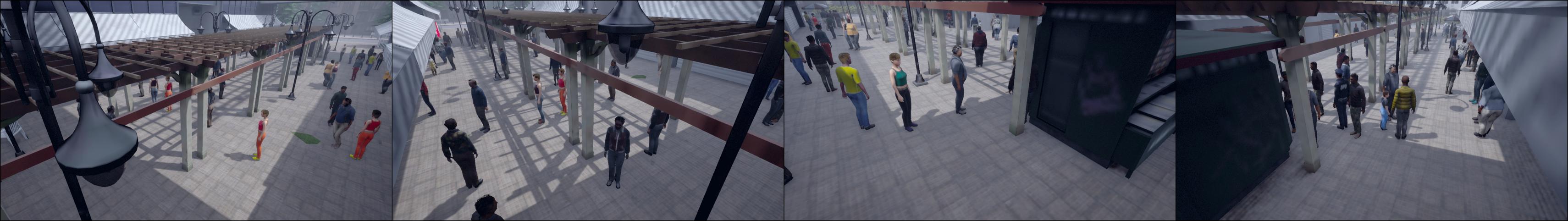} \hfill
        \includegraphics[height=0.107\linewidth]{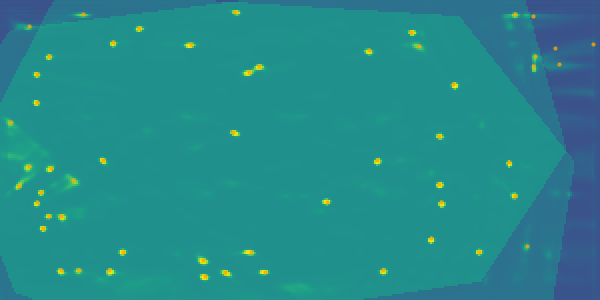} 
        
        \includegraphics[height=0.107\linewidth]{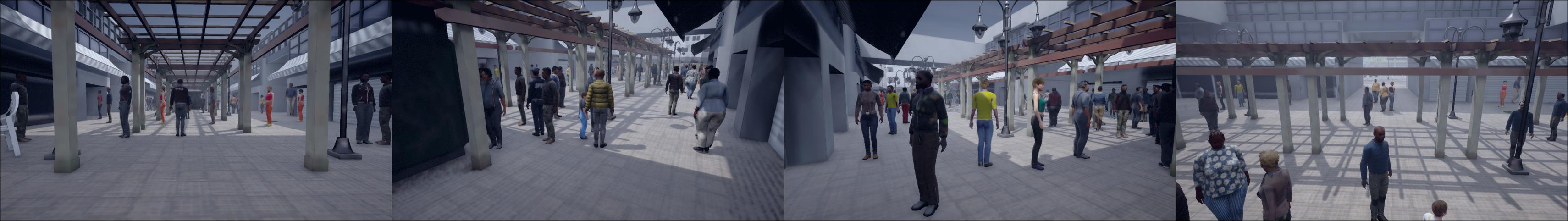} \hfill
        \includegraphics[height=0.107\linewidth]{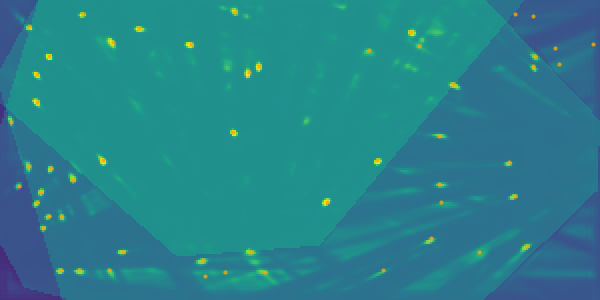} 
        
        \includegraphics[height=0.107\linewidth]{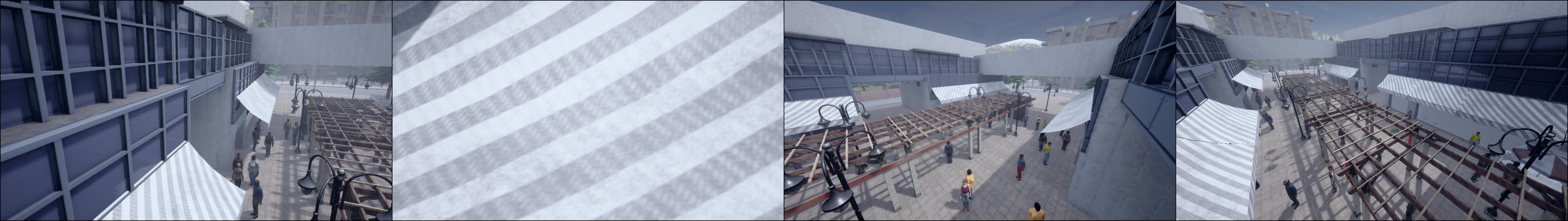} \hfill
        \includegraphics[height=0.107\linewidth]{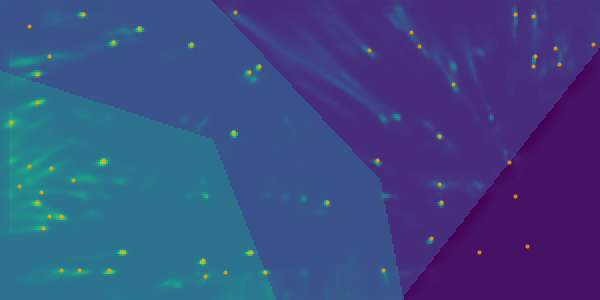} 
        
        \includegraphics[height=0.107\linewidth]{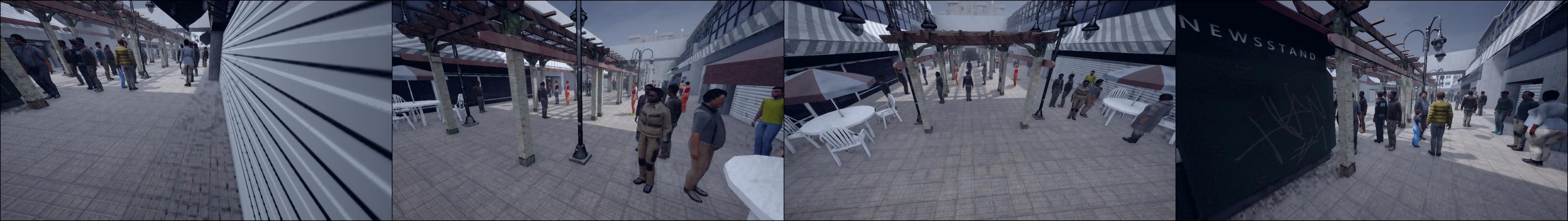} \hfill
        \includegraphics[height=0.107\linewidth]{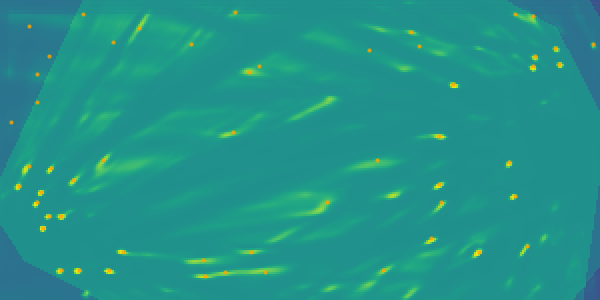} 
        
        \includegraphics[height=0.107\linewidth]{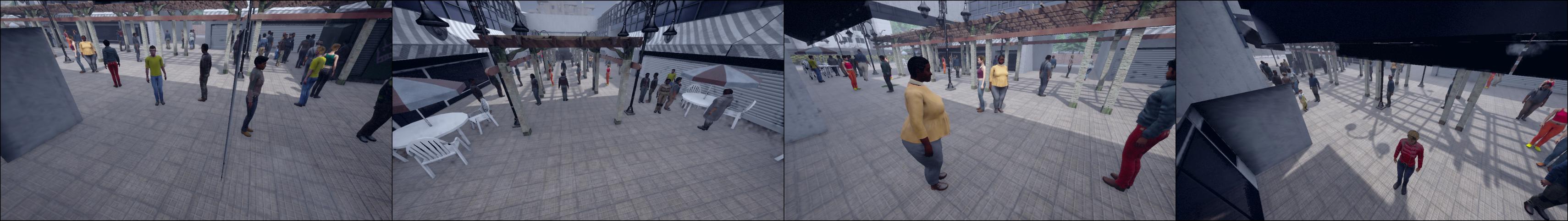} \hfill
        \includegraphics[height=0.107\linewidth]{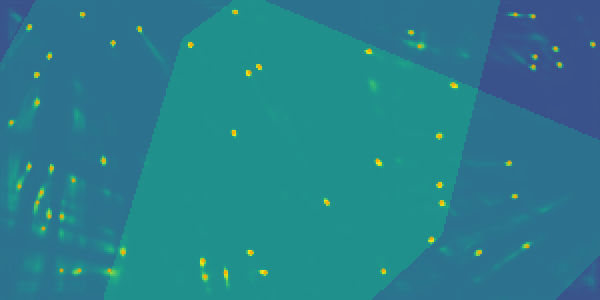} 
    
\caption{
Comparison between camera configurations on \texttt{Town04building}. From top to bottom, we have camera views from Expert 1, Expert 2, Expert 3, random search, max coverage, and the proposed method. We also show FoV coverages and detection results (heatmap overlay) for pedestrians (orange dots). 
}
\end{figure*}

\begin{figure*}
        \includegraphics[height=0.12\linewidth]{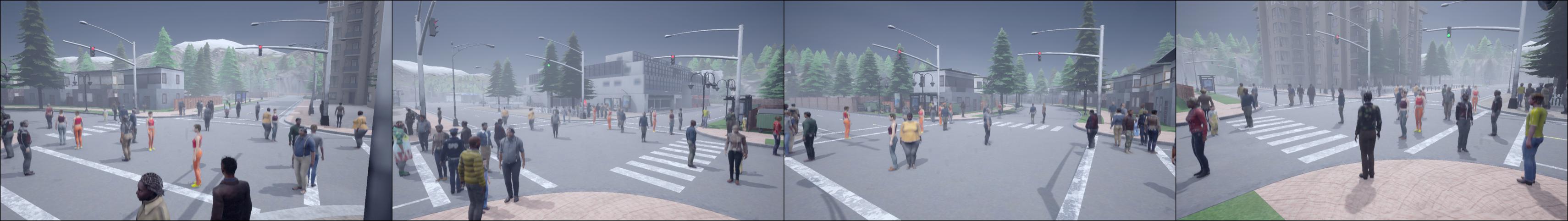} \hfill
        \includegraphics[height=0.12\linewidth]{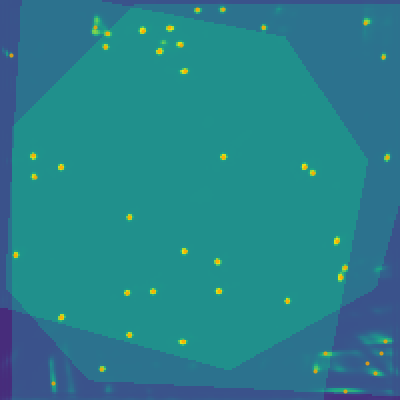} 
        
        \includegraphics[height=0.12\linewidth]{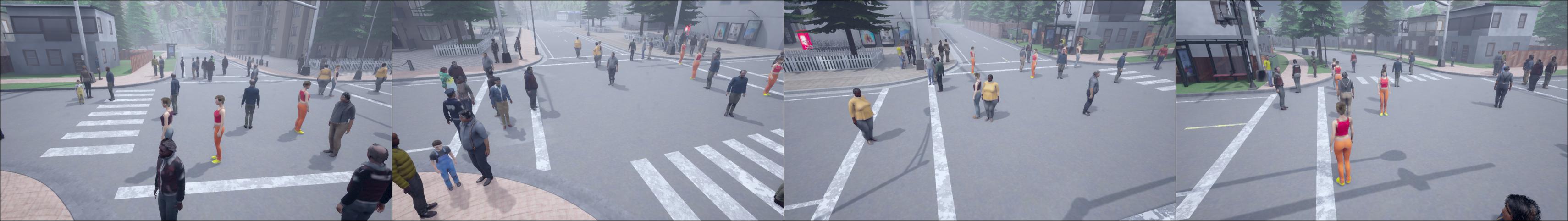} \hfill
        \includegraphics[height=0.12\linewidth]{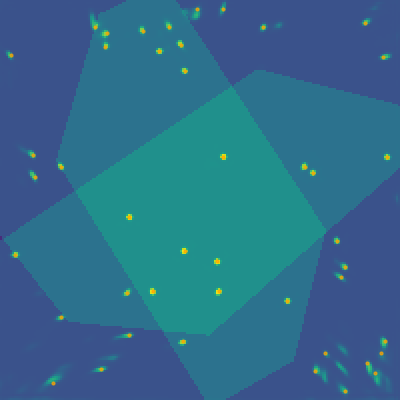} 
        
        \includegraphics[height=0.12\linewidth]{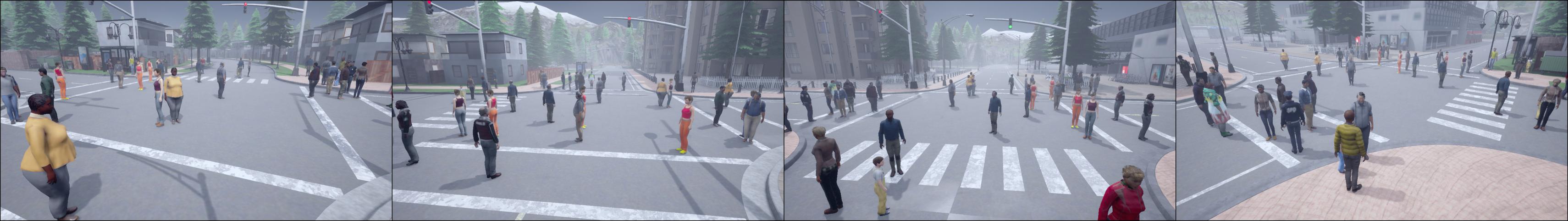} \hfill
        \includegraphics[height=0.12\linewidth]{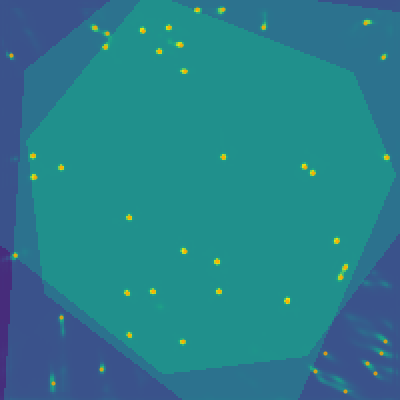} 
        
        \includegraphics[height=0.12\linewidth]{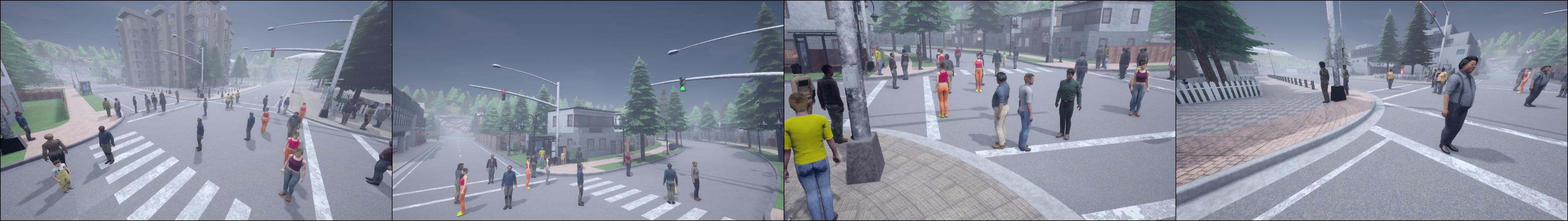} \hfill
        \includegraphics[height=0.12\linewidth]{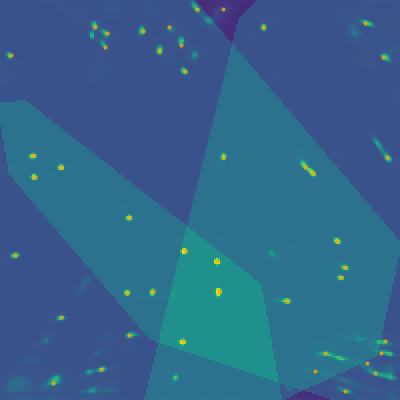} 
        
        \includegraphics[height=0.12\linewidth]{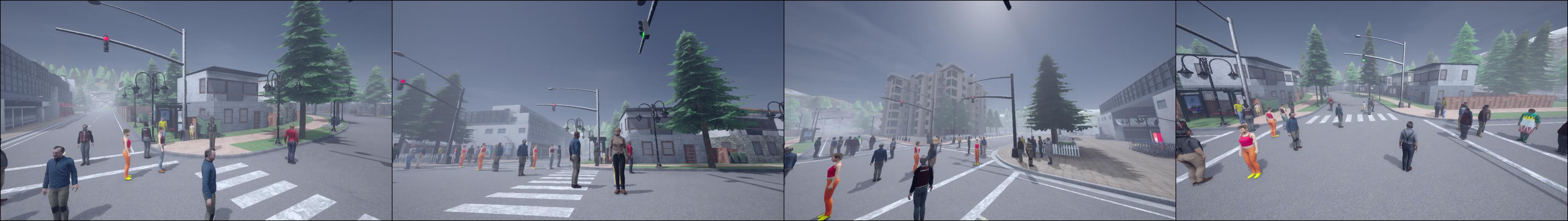} \hfill
        \includegraphics[height=0.12\linewidth]{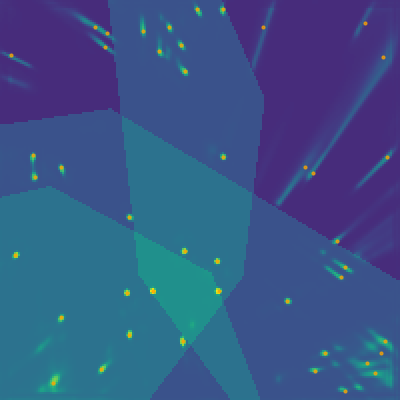} 
        
        \includegraphics[height=0.12\linewidth]{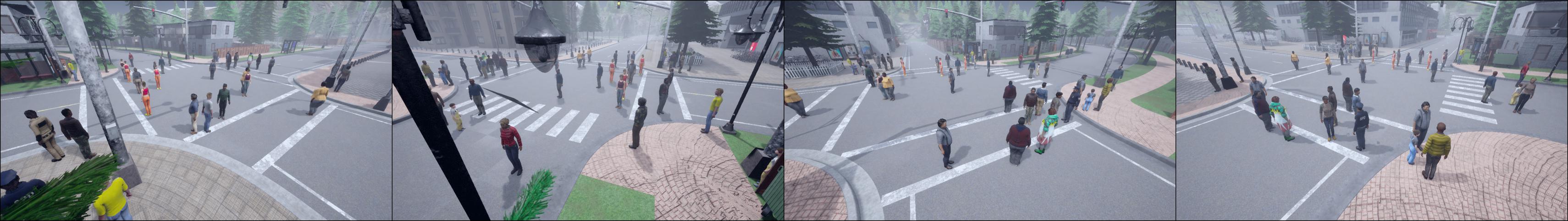} \hfill
        \includegraphics[height=0.12\linewidth]{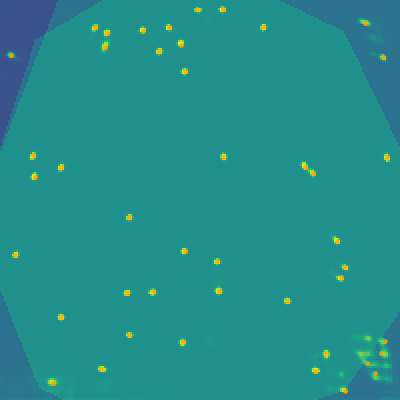} 

\caption{
Comparison between camera configurations on \texttt{Town04crossroad}. From top to bottom, we have camera views from Expert 1, Expert 2, Expert 3, random search, max coverage, and the proposed method. We also show FoV coverages and detection results (heatmap overlay) for pedestrians (orange dots). 
}
\end{figure*}

\begin{figure*}
        \includegraphics[height=0.11\linewidth]{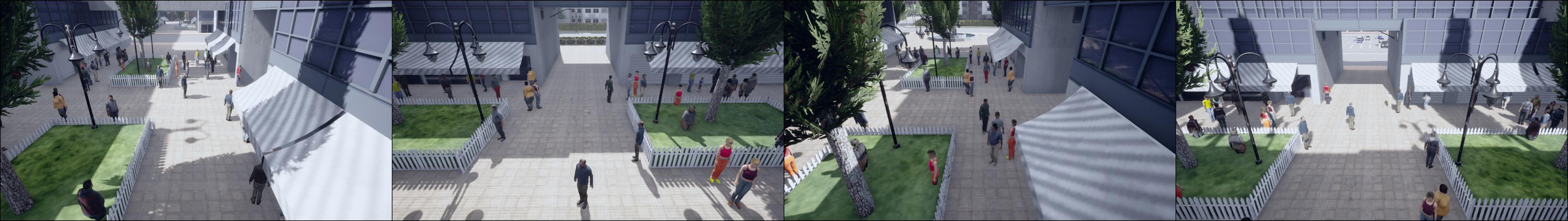} \hfill
        \includegraphics[width=0.11\linewidth,angle=90]{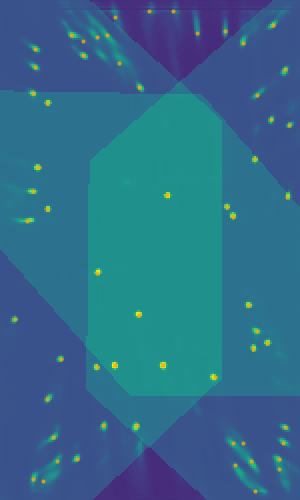} 
        
        \includegraphics[height=0.11\linewidth]{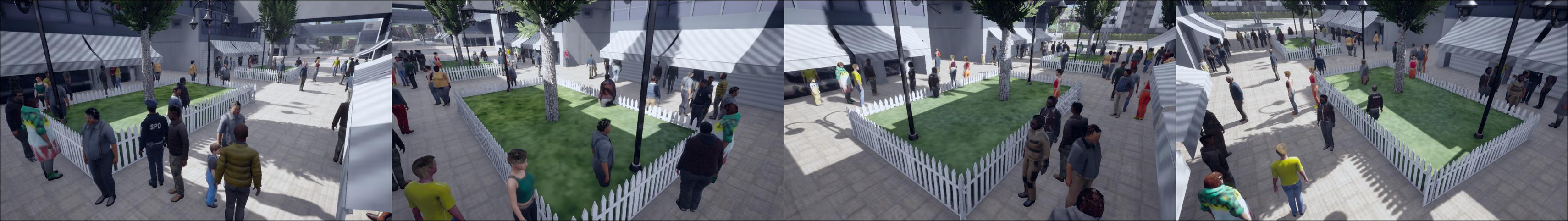} \hfill
        \includegraphics[width=0.11\linewidth,angle=90]{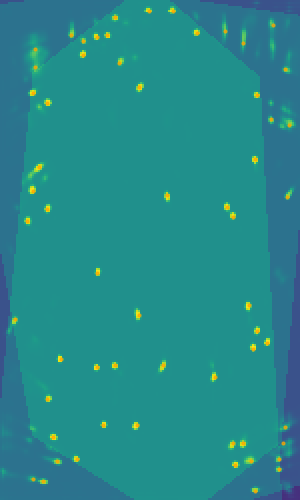} 
        
        \includegraphics[height=0.11\linewidth]{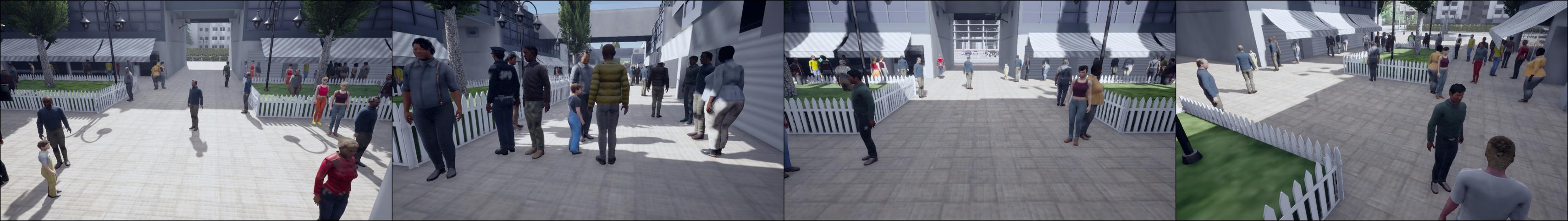} \hfill
        \includegraphics[width=0.11\linewidth,angle=90]{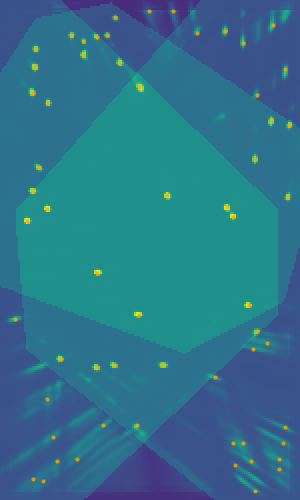} 
        
        \includegraphics[height=0.11\linewidth]{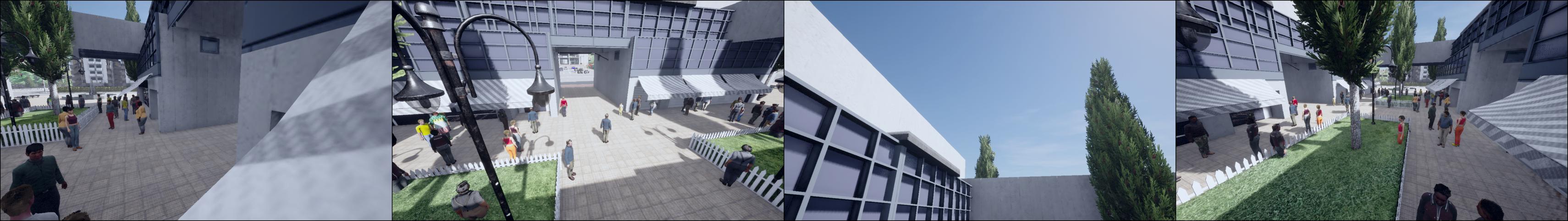} \hfill
        \includegraphics[width=0.11\linewidth,angle=90]{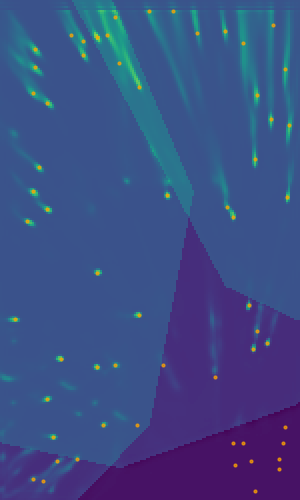} 
        
        \includegraphics[height=0.11\linewidth]{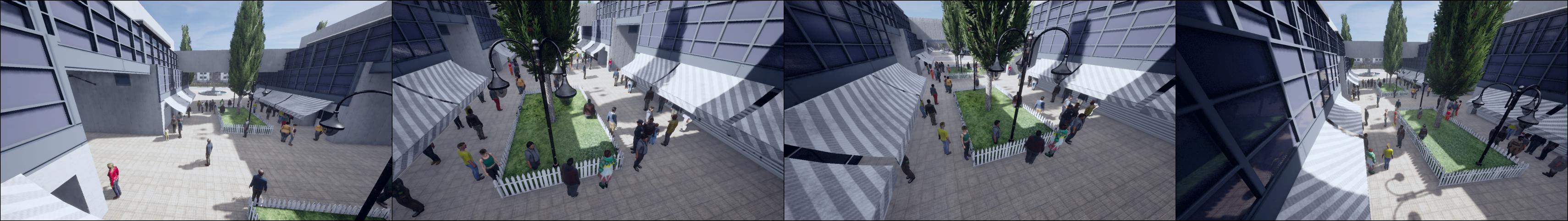} \hfill
        \includegraphics[width=0.11\linewidth,angle=90]{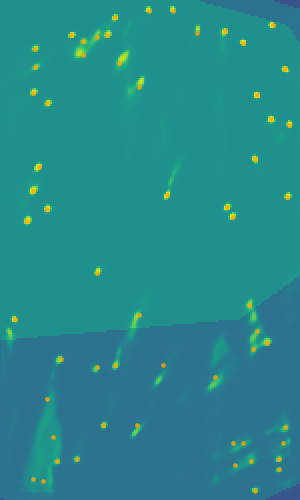} 
        
        \includegraphics[height=0.11\linewidth]{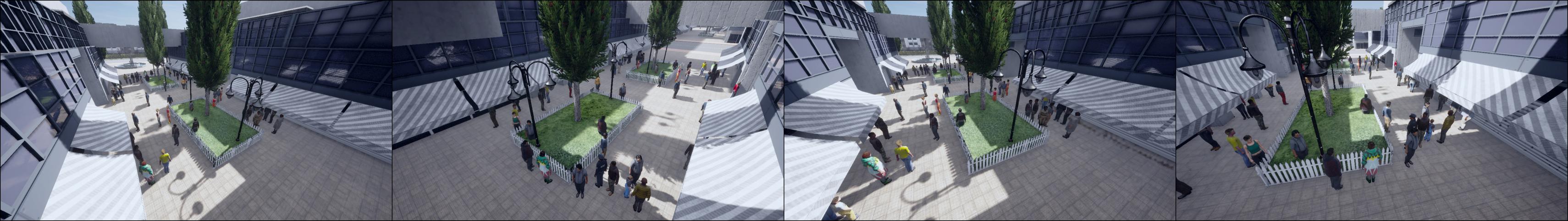} \hfill
        \includegraphics[width=0.11\linewidth,angle=90]{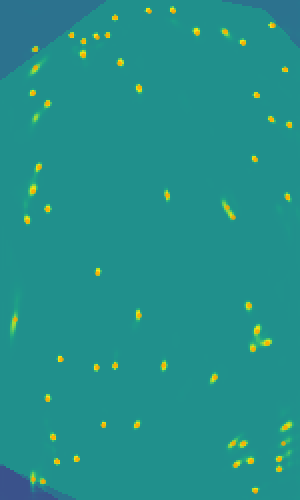} 

\caption{
Comparison between camera configurations on \texttt{Town05building}. From top to bottom, we have camera views from Expert 1, Expert 2, Expert 3, random search, max coverage, and the proposed method. We also show FoV coverages and detection results (heatmap overlay) for pedestrians (orange dots). 
}
\end{figure*}

\begin{figure*}
\centering
        \includegraphics[width=\linewidth]{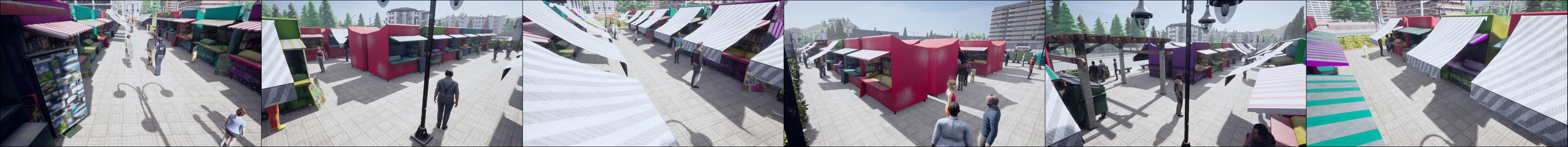} 
        
        \includegraphics[width=0.22\linewidth,angle=0]{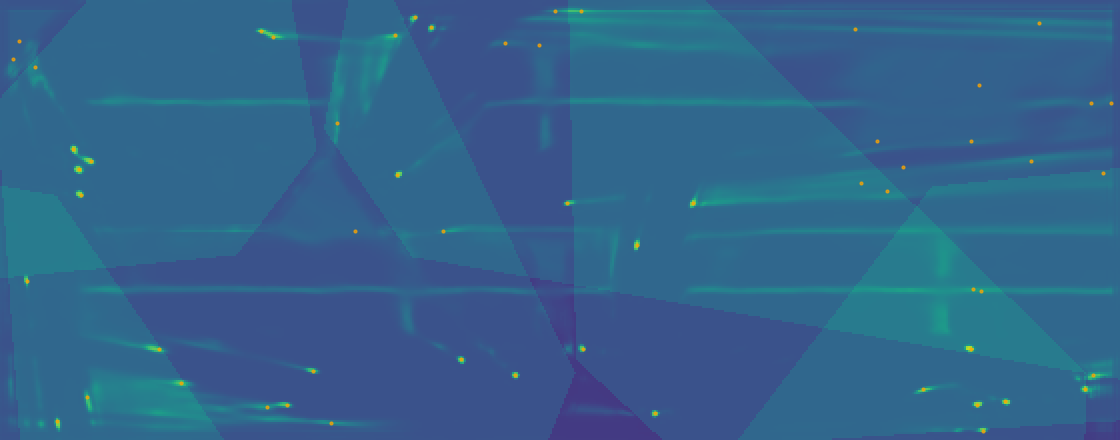} 
        
        \includegraphics[width=\linewidth]{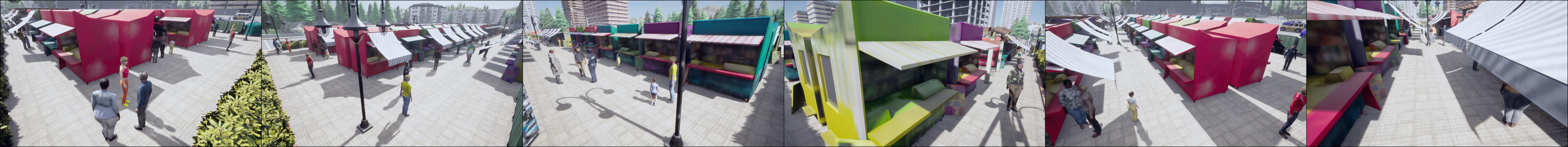} 
        
        \includegraphics[width=0.22\linewidth,angle=0]{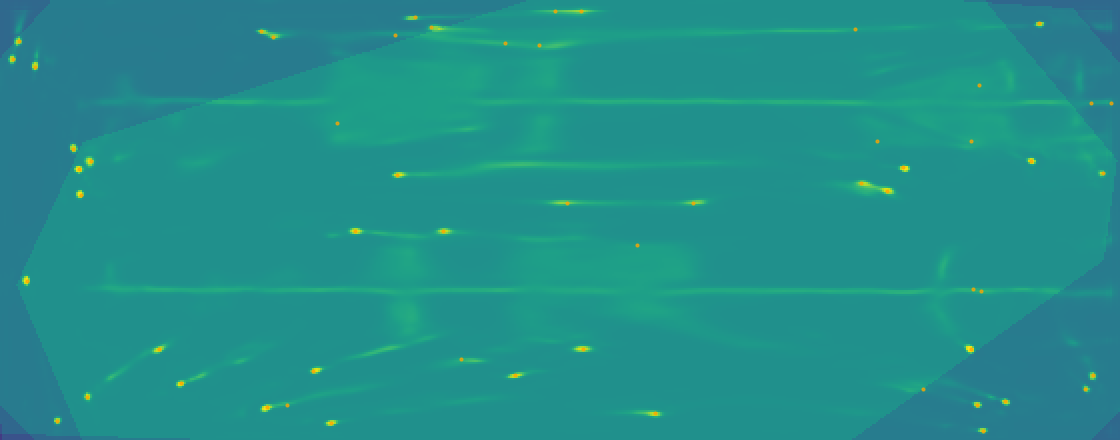} 
        
        \includegraphics[width=\linewidth]{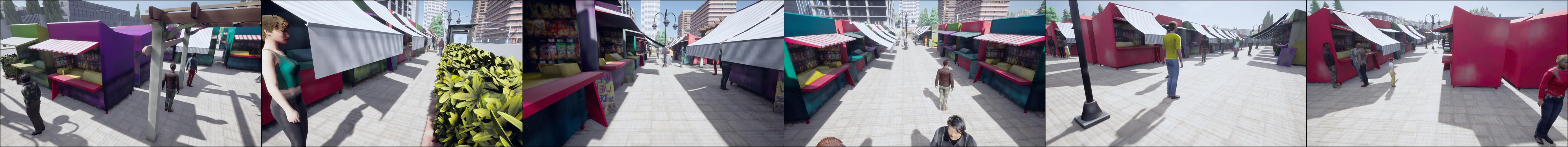} 
        
        \includegraphics[width=0.22\linewidth,angle=0]{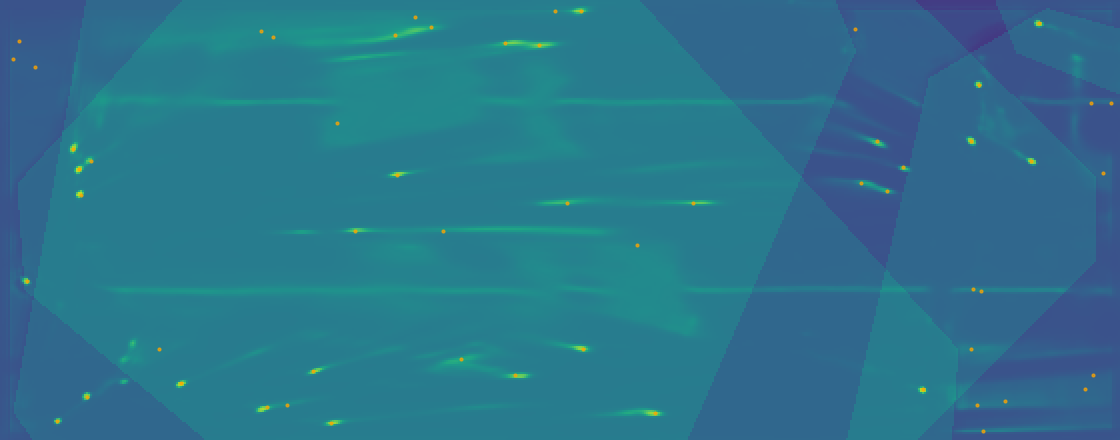} 
        
        \includegraphics[width=\linewidth]{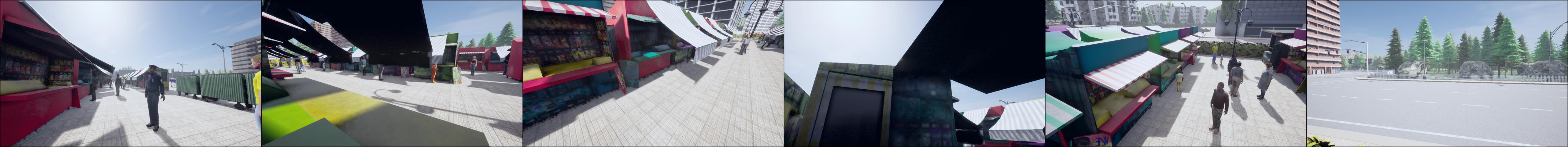} 
        
        \includegraphics[width=0.22\linewidth,angle=0]{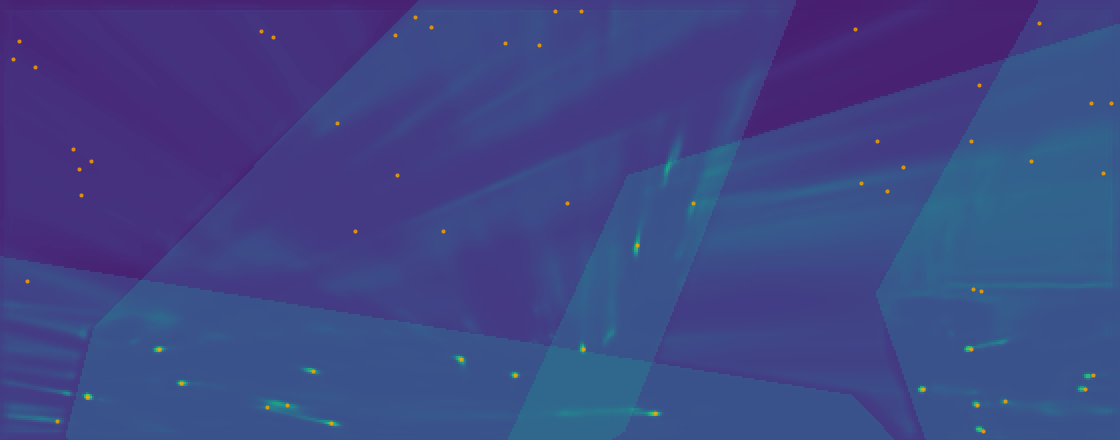} 
        
        \includegraphics[width=\linewidth]{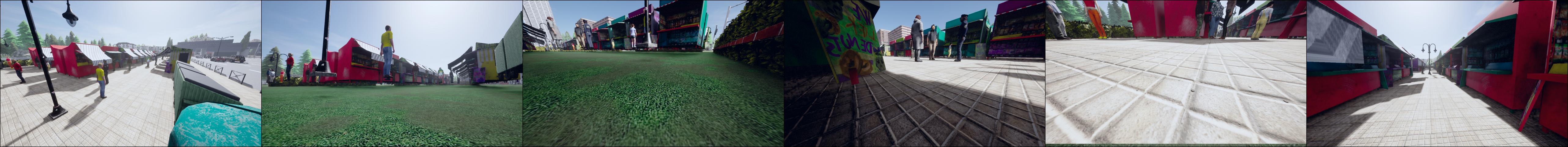}
        
        \includegraphics[width=0.22\linewidth,angle=0]{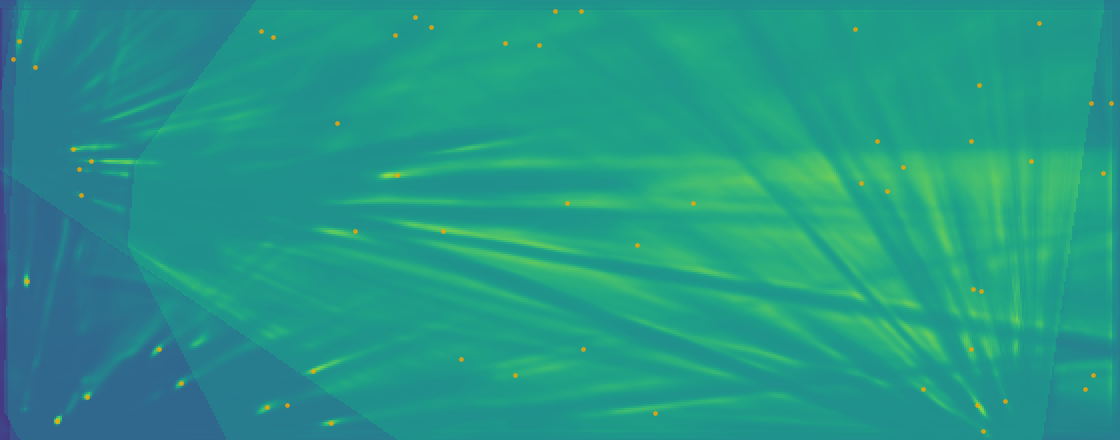} 
        
        \includegraphics[width=\linewidth]{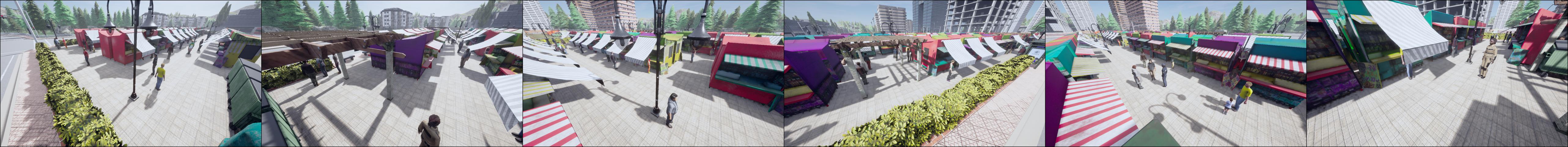}
        
        \includegraphics[width=0.22\linewidth,angle=0]{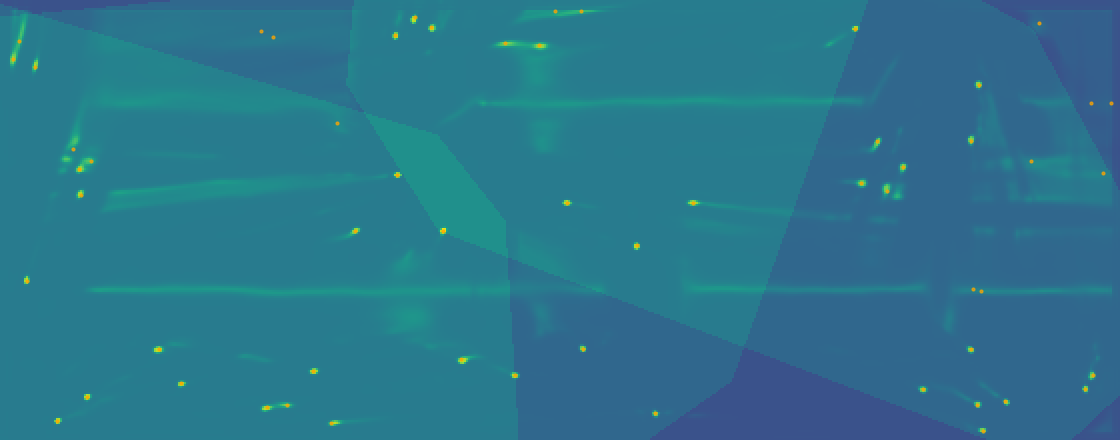} 

\caption{
Comparison between camera configurations on \texttt{Town05market}. From top to bottom, we have camera views from Expert 1, Expert 2, Expert 3, random search, max coverage, and the proposed method. We also show FoV coverages and detection results (heatmap overlay) for pedestrians (orange dots). 
}
\end{figure*}

\begin{figure*}
        \includegraphics[height=0.122\linewidth]{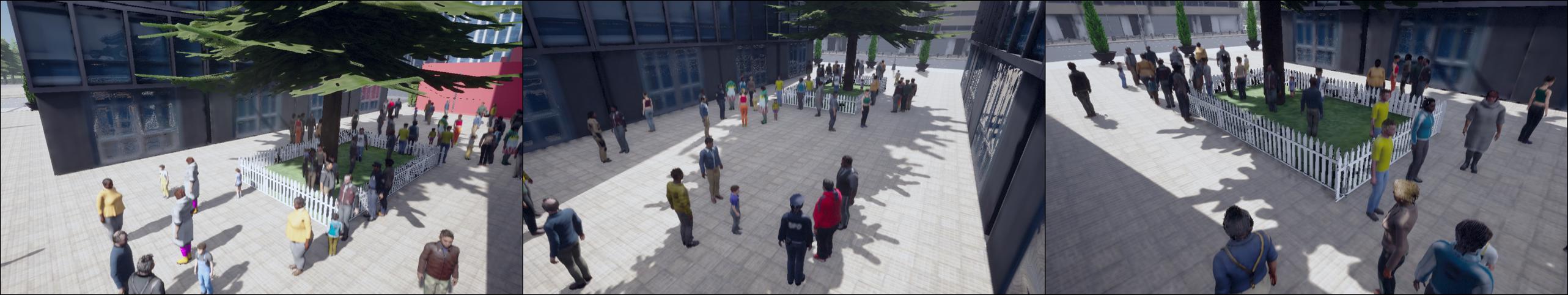} \hfill
        \includegraphics[height=0.122\linewidth]{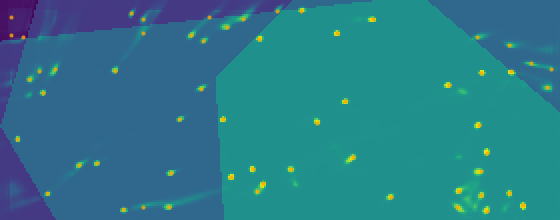} 
        
        \includegraphics[height=0.122\linewidth]{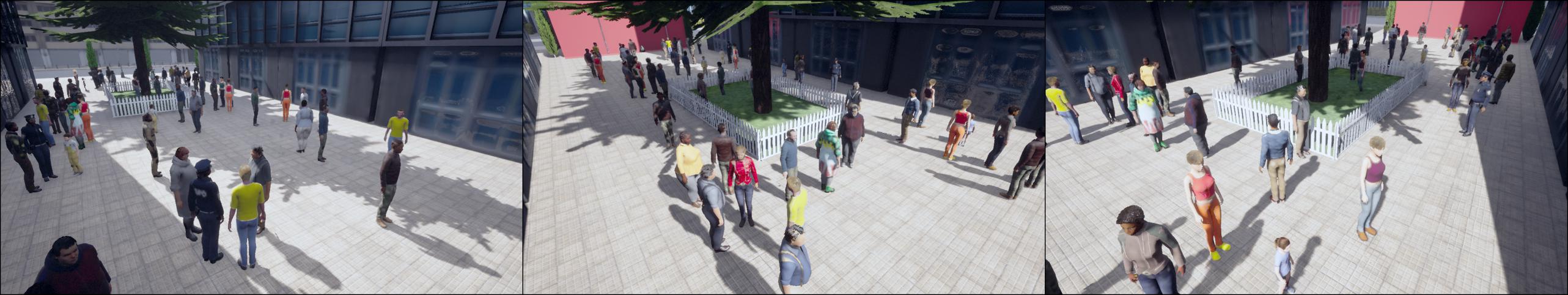} \hfill
        \includegraphics[height=0.122\linewidth]{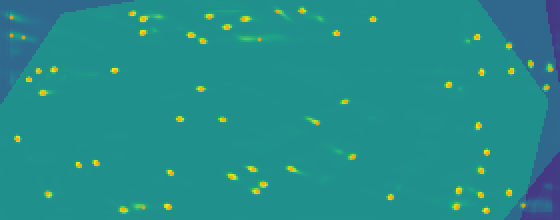} 
        
        \includegraphics[height=0.122\linewidth]{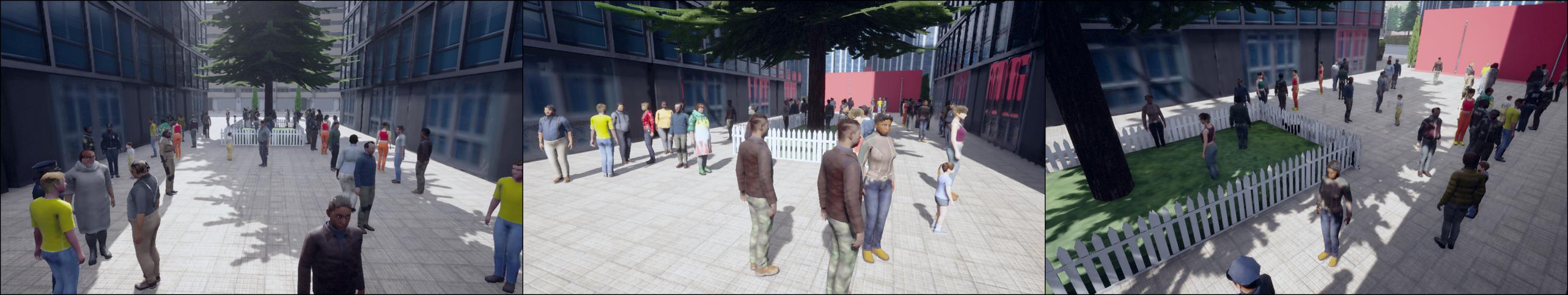} \hfill
        \includegraphics[height=0.122\linewidth]{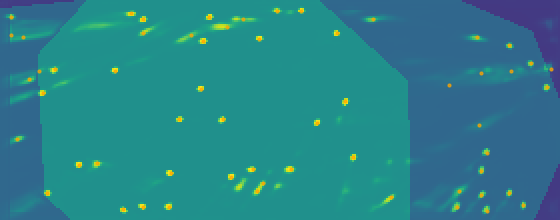} 
        
        \includegraphics[height=0.122\linewidth]{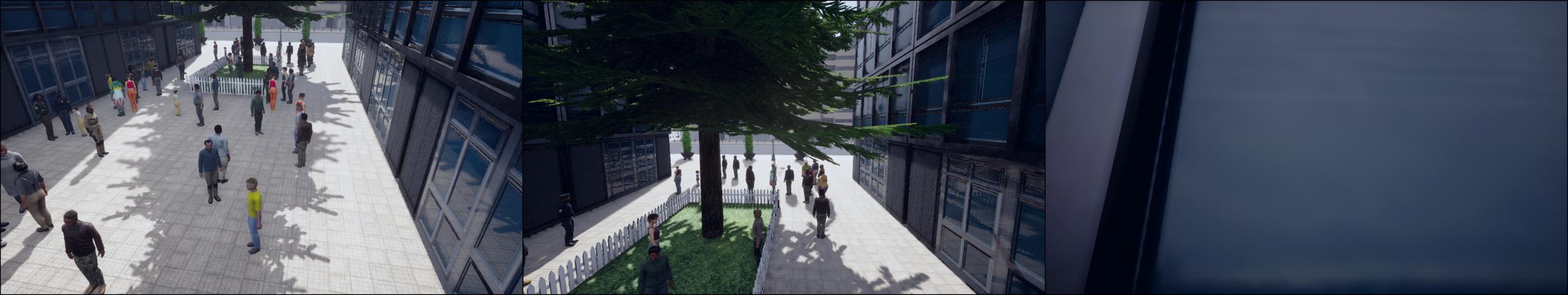} \hfill
        \includegraphics[height=0.122\linewidth]{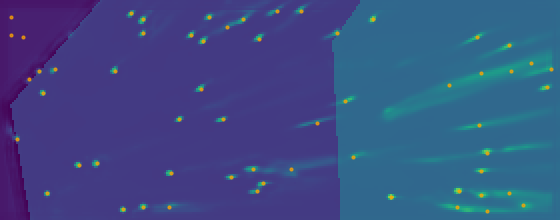} 
        
        \includegraphics[height=0.122\linewidth]{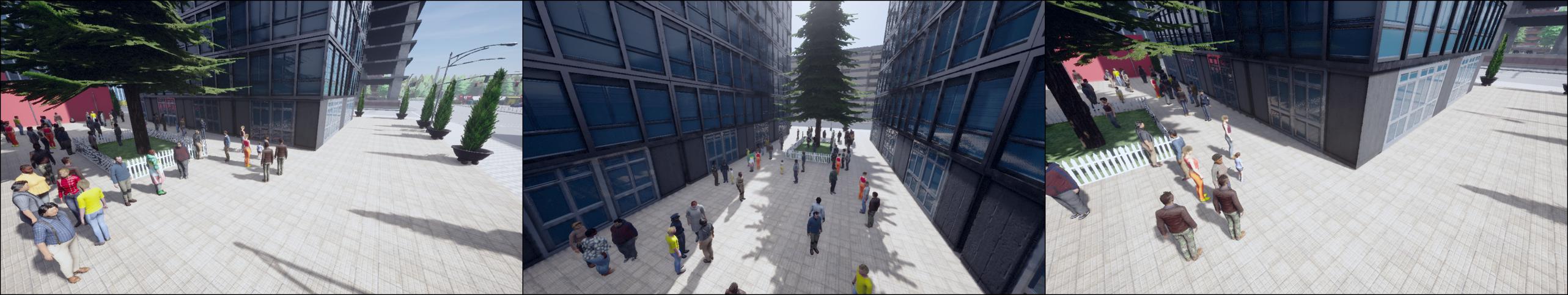} \hfill
        \includegraphics[height=0.122\linewidth]{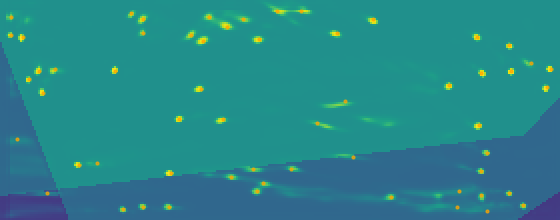} 
        
        \includegraphics[height=0.122\linewidth]{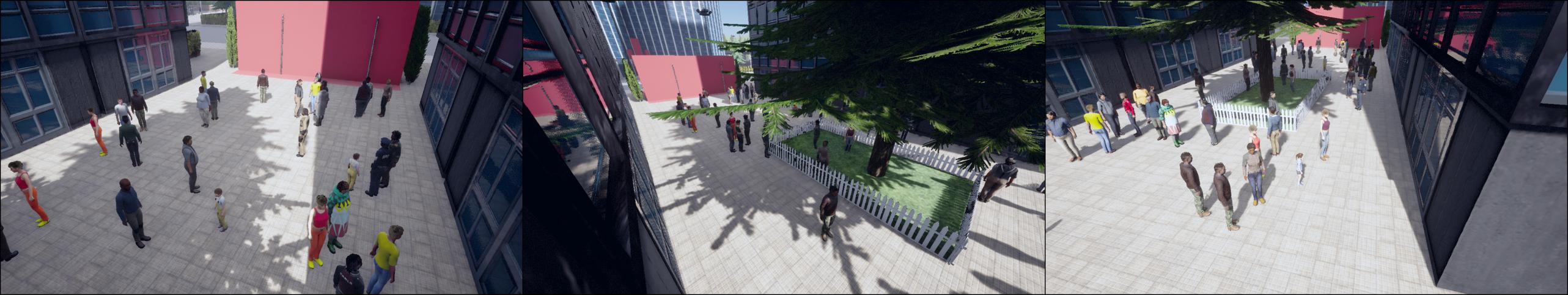} \hfill
        \includegraphics[height=0.122\linewidth]{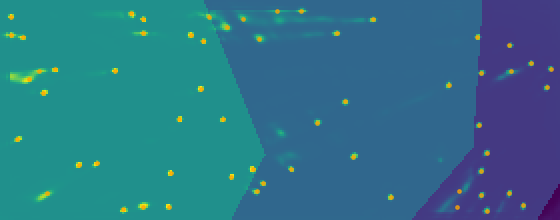} 

\caption{
Comparison between camera configurations on \texttt{Town05skyscraper}. From top to bottom, we have camera views from Expert 1, Expert 2, Expert 3, random search, max coverage, and the proposed method. We also show FoV coverages and detection results (heatmap overlay) for pedestrians (orange dots). 
}
\end{figure*}

\end{document}